\ifdefined\pdfoutput\pdfoutput=1\fi
\ifdefined\pdfminorversion\pdfminorversion=7\fi
\documentclass{article}
\PassOptionsToPackage{table}{xcolor}

\usepackage{microtype}
\usepackage{graphicx}
\usepackage{subfigure}
\usepackage{booktabs} 
\usepackage{hyperref}
\usepackage[preprint]{icml2026}
\setcitestyle{authoryear,round}

\hypersetup{
  pdfauthor={Zhouyang Yu, Shiqi Liu, Zeyu He, Feihong Zhang, Jing Su, Kai Yu, Yang Guan, Shengbo Eben Li},
  pdfsubject={},
  pdfkeywords={}
}
\makeatletter
\renewcommand{\Notice@String}{}
\global\icml@noticeprintedtrue
\makeatother

\usepackage{amsmath}
\usepackage{amssymb}
\usepackage{mathtools}
\usepackage{amsthm}

\usepackage[capitalize,noabbrev]{cleveref}

\theoremstyle{plain}
\newtheorem{theorem}{Theorem}[section]
\newtheorem{proposition}[theorem]{Proposition}

\theoremstyle{definition}

\theoremstyle{remark}

\usepackage{amsfonts}

\usepackage{multirow}
\usepackage[table]{xcolor}
\usepackage{capt-of}
\usepackage{array}
\usepackage{url}
\usepackage{float}
\usepackage{array,xcolor}
\usepackage{algorithm}
\usepackage{algpseudocode}

\usepackage[disable,textsize=tiny]{todonotes}

\icmltitlerunning{Reconciling Conflicting Signals in LLM Post-Training}

\begin{document}

\twocolumn[
\icmltitle{Reconciling Conflicting Signals in LLM Post-Training}

\begin{center}
{\large
Zhouyang Yu\textsuperscript{1} \quad
Shiqi Liu\textsuperscript{1} \quad
Zeyu He\textsuperscript{1} \quad
Feihong Zhang\textsuperscript{1} \\
Jing Su\textsuperscript{2} \quad
Kai Yu\textsuperscript{2} \quad
Yang Guan\textsuperscript{1,$\dagger$} \quad
Shengbo Eben Li\textsuperscript{1,$\dagger$}\\[0.5em]
}
\normalsize \textsuperscript{1}College of AI \& School of Vehicle and Mobility,
Tsinghua University \quad
\textsuperscript{2}Horizon Robotics
\end{center}

\vskip 0.2in
]


\begingroup
\renewcommand{\thefootnote}{\fnsymbol{footnote}}
\footnotetext[2]{Corresponding authors.}
\endgroup

\begin{abstract}
Training a language model on its own responses can exploit three complementary
signals: a response-level verifier outcome, a token-level teacher--student
discrepancy, and a teacher-adaptation signal induced by post-training. However, these signals differ in scale and can prescribe conflicting update directions, making their joint integration challenging.
We introduce \textbf{MOPA} (Multi-Signal Optimization with Probabilistic Advantage), the first three-signal fusion
framework. MOPA maps each raw signal to a comparable probability of the true advantage being positive. Joint Bayesian inference from all observed signals then yields a posterior mean direction that serves as the surrogate advantage. The resulting fusion behavior interpolates between majority voting and weighted linear aggregation. This flexibility mitigates scale dominance while preserving signal strength for conflict resolution.
Experiments on
mathematical and code reasoning show that MOPA improves teacher-normalized
aggregate performance over strong on-policy distillation baselines across
same-size, size-mismatched, and multi-teacher settings.
\end{abstract}

\begin{figure*}[t]
  \centering
  \begin{minipage}[b]{0.19\linewidth}
    \centering
    \includegraphics[width=\linewidth]
      {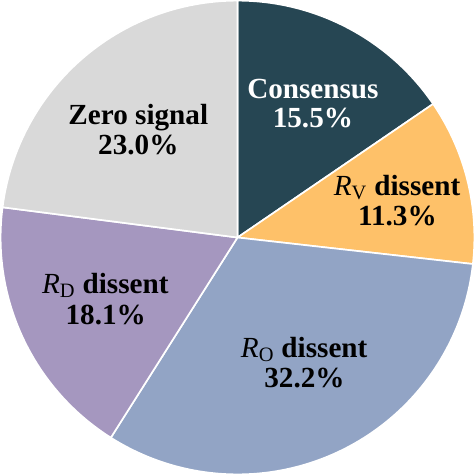}
    \par\vspace{10pt}
    {\bfseries (a)\par}
  \end{minipage}\hfill
  \begin{minipage}[b]{0.23\linewidth}
    \centering
    \includegraphics[width=\linewidth]{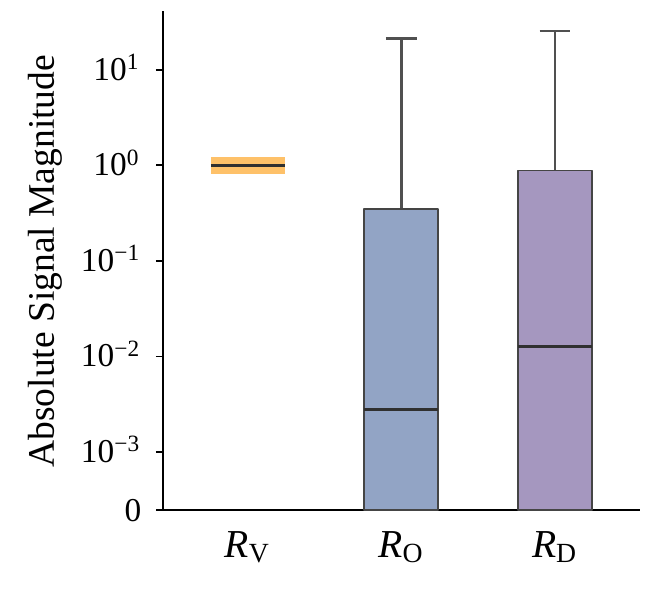}
    \par\vspace{0pt}
    {\bfseries (b)\par}
  \end{minipage}\hfill
  \begin{minipage}[b]{0.248\linewidth}
    \centering
    \includegraphics[width=\linewidth]{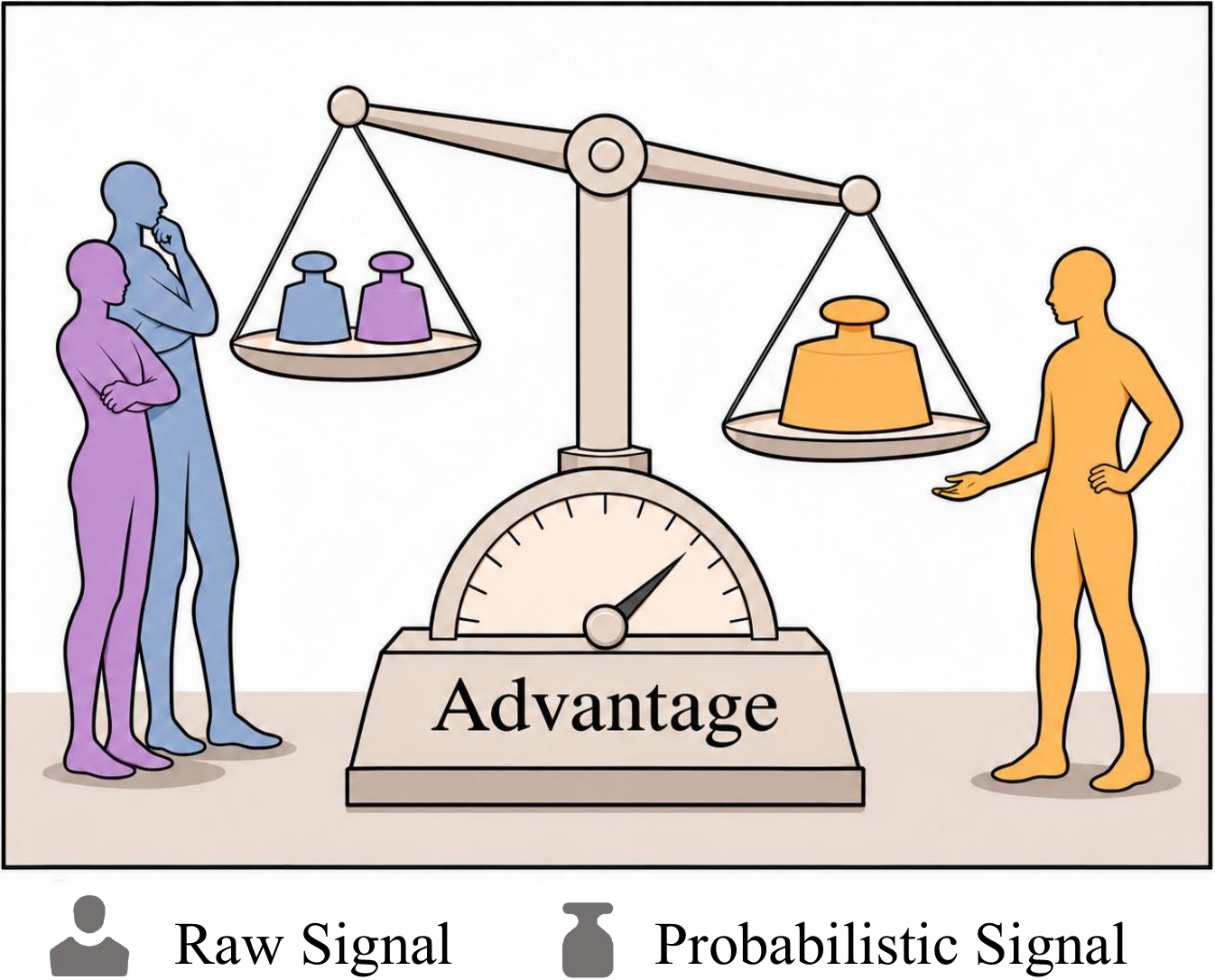}
    \par\vspace{5pt}
    {\bfseries (c)\par}
  \end{minipage}\hfill
  \begin{minipage}[b]{0.25\linewidth}
    \centering
    \includegraphics[width=\linewidth]{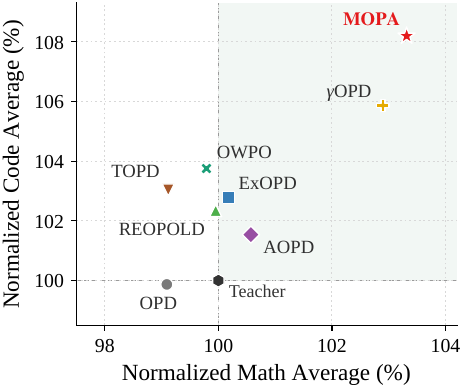}
    \par\vspace{1pt}
    {\bfseries (d)\par}
  \end{minipage}
  
    \caption{\textbf{Learning from conflicting signals with MOPA.}
    (a) Distribution of token-level directional agreement regimes for the three signals: verifier outcome ($R_{\mathrm V}$), teacher--student discrepancy ($R_{\mathrm O}$), and teacher adaptation delta ($R_{\mathrm D}$). Cases include consensus, single-signal dissent, and zero-signal configurations.
    (b) Magnitude disparities among the three raw signals.
    (c) MOPA aligns conflicting signals in probability space to synthesize advantage-direction evidence.
    (d) MOPA achieves the best teacher-normalized mathematical and code reasoning performance in the multi-teacher setting.
    }

  \label{fig:bids-overview-comparison}
\end{figure*}

\section{Introduction}

Post-training has substantially improved large language models' mathematical
reasoning, code generation, and long-horizon problem solving
~\citep{deepseekmath2024,deepseekr12025,
yangQwen3TechnicalReport2025,teamMiMoV2FlashTechnicalReport2026}.
Two prominent approaches are reinforcement learning with verifiable rewards
(RLVR), which uses task outcomes to reinforce student responses, and on-policy
distillation (OPD), which uses a teacher to guide tokens sampled from the
student's policy~\citep{deepseekmath2024,agarwal2024gkd}. If the teacher's
pre-adaptation checkpoint is available, its change during post-training provides a further
token-level signal~\citep{feng2026directopd}. A student rollout can thus
provide three signals for training: the verifier outcome, the
teacher--student discrepancy, and the teacher-adaptation signal.

The three signals answer different questions. The verifier indicates whether
a complete response succeeds, but does not identify which tokens caused that
outcome. The teacher--student discrepancy provides local guidance at each
visited prefix, yet its reliability depends on the compatibility between the
teacher and the student's trajectory. The teacher-adaptation signal isolates
the policy shift induced by the teacher's post-training, providing token-level directions
for potential improvement, but it neither verifies the current response nor
adapts to the student's current state. Consequently, none of the three signals
alone provides both reliable outcome direction and fine-grained, student-aware
credit.

Given these complementary strengths, jointly using all three signals can
provide more complete supervision. Recent studies have begun combining
verifier-, teacher-, and policy-derived supervision
~\citep{yang2026gkd,sun2026reopd,ding2026saf,lin2026opdvr}, but primarily
consider two-signal combinations. Joint fusion is challenging because
the binary verifier reward and token-level log-probability ratios differ
substantially in scale and frequently disagree in direction.
Indeed, 79.9\% of tokens with three nonzero signals exhibit directional
conflict (Figure~\ref{fig:bids-overview-comparison}a).
Direct addition can let signal scale dominate the update, while using
directional agreement alone discards relative signal strength.
A unified framework is therefore needed to align heterogeneous scales
and resolve directional conflicts in token-level updates.

We propose \textbf{MOPA} (Multi-Signal Optimization with Probabilistic Advantage) to jointly integrate the three
signals. Within the policy-gradient framework, each signal serves as a
surrogate advantage, specifying the direction and strength of the update
to each sampled token's probability. MOPA estimates the posterior
expectation of the true advantage direction and uses it as a bounded
surrogate advantage. This places heterogeneous signals in a common
probabilistic space and recasts fusion as posterior inference over
advantage directions given joint signal observations. Under a balanced
prior and conditional independence of signal directions, this posterior
can be computed from directional probabilities conditioned on individual
signals. We design a temperature-controlled formulation of these
single-source probabilities, connecting majority voting and weighted
linear aggregation as limiting update directions.

Our contributions are threefold:
\begin{itemize}
    \item We develop a general Bayesian framework that maps heterogeneous signals with incompatible scales into a common probability space, enabling joint posterior inference of a bounded token-level surrogate advantage.
    \item Within this framework, we instantiate a temperature-controlled probabilistic fusion rule that interpolates between majority voting and weighted linear aggregation, mitigating scale dominance while preserving signal strength for conflict resolution.
    \item We demonstrate the effectiveness of MOPA on mathematical and code reasoning benchmarks, achieving strong performance across same-size, size-mismatched, and multi-teacher settings.
\end{itemize}

\section{Related Work}

\subsection{Single-Signal Post-Training}

\textbf{(1) Verifier-based reinforcement learning.}
GRPO normalizes verifiable rewards within each rollout group and broadcasts the
resulting response-level advantage to every token~\citep{deepseekmath2024}.
DAPO extends this objective with decoupled clipping, dynamic sampling, and
token-level loss aggregation~\citep{yu2025dapo}. Entropy- and probability-based
reweighting further distinguishes reasoning forks from spurious gradients
~\citep{wang2025entropy,liu2026stapo}; all use only verifier feedback and
student-policy statistics.

\textbf{(2) Teacher--student on-policy distillation.}
OPD evaluates student-generated trajectories with a teacher, providing dense
token-level supervision on visited states~\citep{agarwal2024gkd}. Refinements
address estimator instability and teacher--student mismatch
~\citep{fu2026revisiting,li2026rethinking}, but still use only their current
policy discrepancy.

\textbf{(3) Teacher-adaptation distillation.}
Direct-OPD transfers the policy shift between a post-trained teacher and its pre-adaptation
reference as a dense implicit reward on student-generated trajectories, enabling
weak-to-strong transfer without imitating the teacher's final policy
~\citep{feng2026directopd}.

\subsection{Multi-Signal Post-Training}

\textbf{(1) Integration of teacher--student discrepancy and teacher-adaptation signals.}
G-OPD combines the teacher--student discrepancy signal with a scaled
teacher-adaptation residual~\citep{yang2026gkd}. To constrain updates driven by the teacher-adaptation signal,
OPD$^2$ retains this signal only where its centered advantage agrees in
sign with the OPD advantage~\citep{heo2026opd2}. REOPD instead controls
extrapolation along this shift using student--teacher compatibility and
a batch-level reliability budget~\citep{sun2026reopd}. All combine two teacher-derived signals without
independently verifying task success.

\begin{figure*}[t]
  \centering
  \includegraphics[width=\linewidth]{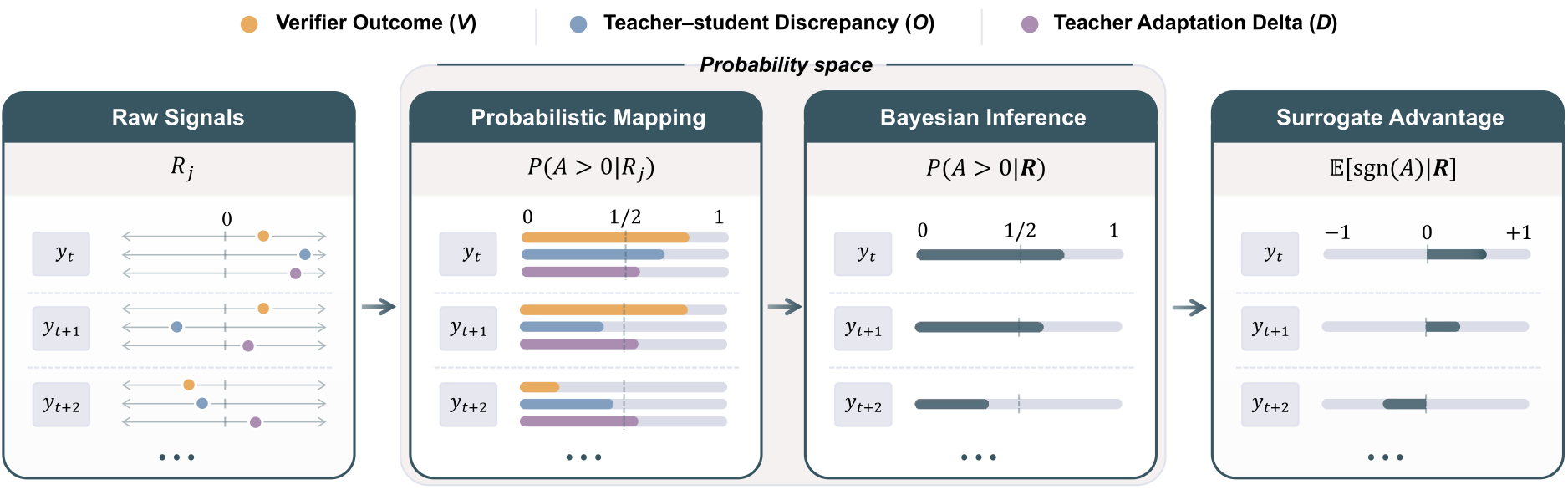}
  \caption{\textbf{General framework of MOPA.} Raw signals are probabilistically
  aligned and combined through Bayesian aggregation to infer the posterior mean direction used as a surrogate
  advantage.}
  \label{fig:beta-method}
\end{figure*}

\textbf{(2) Integration of verifier and teacher--student discrepancy signals.}
This family includes additive fusion with magnitude or schedule control
~\citep{ding2026saf}, verifier-anchored magnitude modulation
~\citep{yang2026rlsd,yang2026owpo}, agreement-based gating or rectification
~\citep{akhondzadeh2026rgopd,xu2026sgopd,lin2026opdvr}, and trajectory-level
calibration of teacher scores~\citep{hou2026uniopd,zhang2026gcopd}. These methods address scale imbalance or directional conflict through
method-specific weighting, gating, or calibration, typically for two
sources. Extending them to three signals requires jointly handling
heterogeneous scales and conflicting directions without fixing one
source's priority. MOPA provides a unified probabilistic framework
that combines evidence from all three sources to infer the update direction.

\section{Preliminaries}

For a prompt $x\sim\mathcal D$ drawn from the prompt distribution $\mathcal D$, let
$y=(y_{1},\ldots,y_{T})$ denote a response of length $T$ and
$h_{t}=(x,y_{<t})$ the prefix preceding token $y_{t}$. We use
$\pi_\theta$ for the student policy with parameters $\theta$, $\pi_T$ for
the fixed post-trained teacher, and $\pi_{T_b}$ for its pre-adaptation checkpoint. The signals $R_{\mathrm V}$,
$R_{\mathrm O,t}$, and $R_{\mathrm D,t}$ denote the verifier outcome,
teacher--student discrepancy, and teacher-adaptation signal, respectively.
For compactness, $\mathbb E$ below averages over
$x\sim\mathcal D$ and $y\sim\pi_\theta(\cdot\mid x)$.

\paragraph{Reinforcement learning with verifiable reward.}
RLVR samples $y\sim\pi_\theta(\cdot\mid x)$ and maximizes the
response-level reward supplied by a verifier~\citep{deepseekmath2024,deepseekr12025}.
We use $R_{\mathrm V}\in\{-1,1\}$ directly for every response token,
without a group-relative transformation. The loss and its policy gradient are
\begin{equation}
\mathcal L_{\mathrm{RLVR}}(\theta)
=-\mathbb E\left[R_{\mathrm V}\right],
\label{eq:rl-loss}
\end{equation}
\begin{equation}
\nabla_\theta\mathcal L_{\mathrm{RLVR}}
=-\mathbb E\!\left[\sum_{t=1}^{T}R_{\mathrm V}\nabla_\theta\log\pi_\theta(y_{t}\mid h_{t})\right].
\label{eq:rl-gradient}
\end{equation}

\paragraph{On-policy distillation.}
OPD trains on student-generated responses by
minimizing the reverse KL divergence~\citep{agarwal2024gkd,yang2026gkd}:
\begin{equation}
\mathcal L_{\mathrm{OPD}}(\theta)
=\mathbb E\!\left[D_{\mathrm{KL}}\!\left(
\pi_\theta(\cdot\mid x)\,\|\,\pi_T(\cdot\mid x)
\right)\right].
\label{eq:opd-loss}
\end{equation}
Using the common next-token approximation, its gradient is
\begin{equation}
\nabla_\theta\mathcal L_{\mathrm{OPD}}
\approx-\mathbb E\!\left[\sum_{t=1}^{T}R_{\mathrm O,t}\nabla_\theta\log\pi_\theta(y_{t}\mid h_{t})\right],
\label{eq:opd-gradient}
\end{equation}
where $R_{\mathrm O,t}=\log\pi_T(y_{t}\mid h_{t})
-\log\pi_\theta(y_{t}\mid h_{t})$ supplies dense token-level supervision
by measuring the teacher--student log-probability discrepancy.

\paragraph{Generalized on-policy distillation.}
G-OPD augments the OPD signal with a teacher-adaptation signal. Using
$\pi_{T_b}$ as the reference, its loss can be written as~\citep{yang2026gkd}
\begin{equation}
\mathcal L_{\mathrm{G\text{-}OPD}}(\theta)
=-\mathbb E\!\left[\sum_{t=1}^{T}\left(R_{\mathrm O,t}
+(\lambda-1)R_{\mathrm D,t}\right)\right].
\label{eq:gopd-equivalent-loss}
\end{equation}
Here, $R_{\mathrm D,t}=\log\pi_T(y_{t}\mid h_{t})
-\log\pi_{T_b}(y_{t}\mid h_{t})$ captures the policy shift induced by the teacher's post-training.
The choice $\lambda>1$ extrapolates beyond standard OPD and controls extrapolation strength.

\section{Method}

\begin{figure*}[t]
  \centering
  \includegraphics[width=0.98\linewidth]{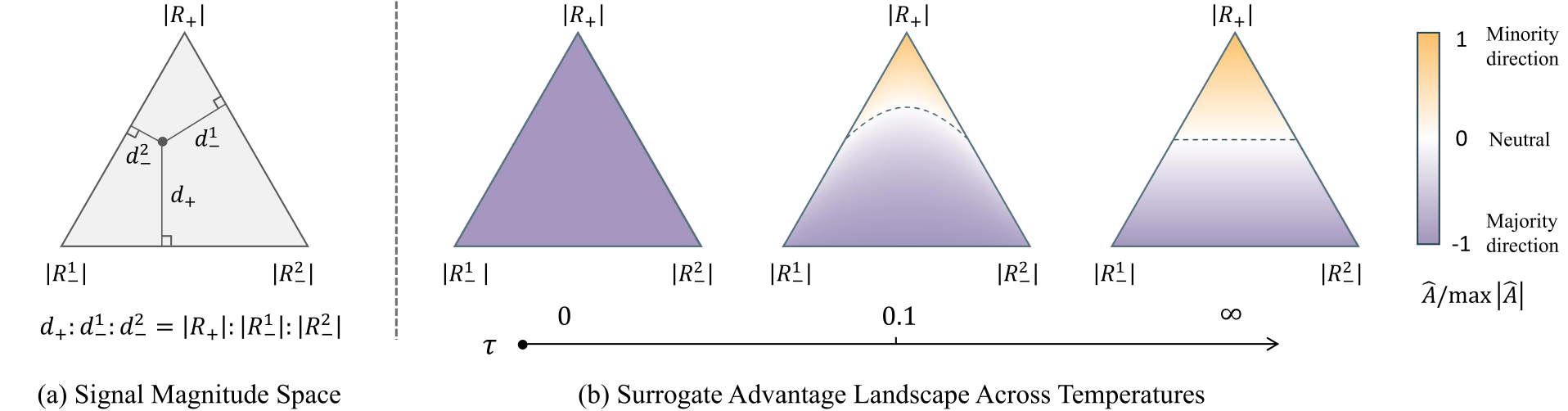}
  \caption{\textbf{Temperature controls the fusion of conflicting signals.}
(a) Barycentric coordinates of signal magnitudes, where the perpendicular distances to the opposite edges encode their relative magnitudes.
(b) Surrogate advantage landscapes for a representative conflicting configuration with one positive signal ($R_+$) and two negative signals ($R^1_{-},R^2_{-}$), using $w_j=1$ for all sources. Color indicates the normalized surrogate advantage, with yellow and purple representing the positive (minority) and negative (majority) directions, respectively. As $\tau$ increases, the fusion rule transitions from majority voting through magnitude-sensitive fusion to linear aggregation in the limit $\tau\to\infty$.}
  \label{fig:bids-temperature}
\end{figure*}

\subsection{General Probabilistic Advantage Synthesis}
\label{sec:fusion-theory}

Within the policy-gradient framework, the true token advantage $A_{t}$
measures how a sampled token changes expected reward relative to the
policy's average action at the same prefix. It determines the direction
and strength of the token's contribution to the policy gradient \citep{li2023reinforcement, guan2021direct}:
\begin{equation}
\nabla_\theta\mathcal L_{\mathrm{PG}}(\theta)
=-\mathbb E\!\left[
\sum_{t=1}^{T}A_{t}\,
\nabla_\theta\log\pi_\theta(y_{t}\mid h_{t})
\right].
\label{eq:advantage-policy-gradient}
\end{equation}
The conditional-expectation definition of $A_{t}$ and the derivation
are given in Appendix~\ref{app:policy-gradient-advantage}.
The verifier, teacher--student discrepancy, and teacher-adaptation signals can help estimate its
direction, but their magnitudes have different scales and are difficult to
align directly. To align these signals probabilistically, we propose
\textbf{MOPA} (Multi-Signal Optimization with Probabilistic Advantage), a framework that infers the advantage direction from
multiple supervision signals and uses its posterior mean to guide policy
updates, as illustrated in Figure~\ref{fig:beta-method}.

For the remainder of this section, we omit the token index $t$. Thus $A$ denotes the true token advantage, assumed nonzero
almost surely, and
$\boldsymbol R=(R_{\mathrm V},R_{\mathrm O},R_{\mathrm D})$ the observed
signals, with $R_{\mathrm V}$ shared by all tokens in a response.
Agreement means $A\cdot R_j>0$ for $j\in\{\mathrm V,\mathrm O,\mathrm D\}$.
Throughout the following probabilistic model, all probabilities are
conditioned on the observed magnitudes.
For each nonzero signal, define its directional reliability as the
probability that its sign agrees with the true advantage:
\begin{equation}
p_j(R_j):=P(A\cdot R_j>0\mid R_j).
\label{eq:source-likelihood}
\end{equation}
The corresponding single-source posterior over advantage directions is
\begingroup
\setlength{\abovedisplayskip}{4pt plus 1pt minus 1pt}
\setlength{\belowdisplayskip}{4pt plus 1pt minus 1pt}
\begin{equation}
P(\operatorname{sgn}(A)=s\mid R_j)
=
\begin{cases}
p_j(R_j), & s=\operatorname{sgn}(R_j),\\
1-p_j(R_j), & s=-\operatorname{sgn}(R_j).
\end{cases}
\label{eq:single-source-direction}
\end{equation}
\endgroup
Thus, a reliable negative signal favors a negative advantage,
rather than a positive update.

\begin{proposition}[Bayesian Linear Accumulation of Directional Evidence]
\label{prop:linear-evidence}
Under Equation~\eqref{eq:source-likelihood}, the balanced conditional prior
$P(A>0)=P(A<0)=1/2$, and
conditional independence of the signal directions given the advantage
direction and observed magnitudes, Bayes' rule gives
the posterior log odds as a linear sum
of signed evidence contributions:
\begin{equation}
L:=\log\frac{P(A>0\mid\boldsymbol R)}{P(A<0\mid\boldsymbol R)}
=\sum_{j}
\operatorname{sgn}(R_j)\,e_j(R_j),
\label{eq:general-evidence-sum}
\end{equation}
where the evidence associated with source $j$'s reliability is
\begin{equation}
e_j(R_j):=\log p_j(R_j)-\log\bigl(1-p_j(R_j)\bigr).
\label{eq:source-evidence}
\end{equation}
\end{proposition}
The proof is given in Appendix~\ref{app:linear-evidence-proof}.
Each signed contribution measures how one signal changes the odds between
the two advantage directions. We model informative sources as better than
chance, $p_j>1/2$, so $e_j>0$ and the sign of each contribution is
determined by the observed signal. 

We define the surrogate advantage directly as the posterior expectation
of the true advantage direction:
\begin{equation}
\widehat A:=\mathbb E[\operatorname{sgn}(A)\mid\boldsymbol R]
=\tanh(L/2).
\label{eq:posterior-mean-direction}
\end{equation}
Its derivation is given in Appendix~\ref{app:posterior-direction-proof}.
The posterior mean minimizes
posterior squared error when estimating $\operatorname{sgn}(A)$, while its
sign selects the more probable update direction. Its magnitude encodes
directional confidence: ambiguous evidence weakens the update, and reliable agreement strengthens the update within a
fixed bound $|\widehat A|\leq1$. Together, logarithmic evidence
compression and the bounded posterior mean naturally attenuate the
influence of anomalous signals with extreme magnitudes, preventing them
from producing arbitrarily large surrogate advantages during training.

To train the student, we use
$\widehat A_{t}$ as the weight in the policy-gradient update. For prompts $x\sim\mathcal D$ and responses
$y\sim\pi_\theta(\cdot\mid x)$, the corresponding loss gradient is
\begin{equation}
\nabla_\theta\mathcal L_{\mathrm{MOPA}}(\theta)\\
=-\mathbb E\!\left[
\sum_{t=1}^{T}\widehat A_{t}\,
\nabla_\theta\log\pi_\theta(y_{t}\mid h_{t})
\right].
\label{eq:beta-gradient}
\end{equation}

\subsection{Tempered Probabilistic Advantage Synthesis}
\label{sec:evidence-fusion}

\begin{table*}[t] \centering
\caption{\textbf{Mathematical reasoning under same-size and size-mismatched distillation.} The average scores of individual benchmarks ($\dagger$) are normalized by the corresponding teacher model's performance on each task before averaging.}
\label{tab:bids-math}
\begingroup
\setlength{\tabcolsep}{0pt}
\newcolumntype{I}{!{\color{black!20}\vrule width 0.4pt}}
\newcolumntype{C}[1]{>{\centering\arraybackslash}m{#1}}
\begin{tabular}{l<{\hspace{15pt}}I C{0.08\textwidth} C{0.08\textwidth}I C{0.08\textwidth} C{0.08\textwidth}I C{0.08\textwidth} C{0.08\textwidth}I C{0.08\textwidth} C{0.08\textwidth}I C{0.08\textwidth} C{0.08\textwidth}}
\toprule
\multirow{2}{*}{Model} & \multicolumn{2}{cI}{AIME24} & \multicolumn{2}{cI}{AIME25} & \multicolumn{2}{cI}{AMC23} & \multicolumn{2}{cI}{MinervaMath} & \multicolumn{2}{c}{$\text{Average}^\dagger$} \\
\cmidrule(lr){2-3}\cmidrule(lr){4-5}\cmidrule(lr){6-7} \cmidrule(lr){8-9}\cmidrule(lr){10-11}
& Acc. & Pass & Acc. & Pass & Acc. & Pass & Acc. & Pass & Acc. & Pass \\
\midrule
\multicolumn{11}{l}{\textit{Teacher: Qwen3-4B-Math}} \\
Teacher & 57.40 & 76.67 & 51.67 & 66.67 & 93.52 & 97.50 & 27.85 & 33.46 & 100.00 & 100.00 \\
\midrule
\multicolumn{11}{l}{\textit{Student: Qwen3-4B}} \\
Base & 22.60 & 60.00 & 20.83 & 33.33 & 60.16 & 92.50 & 20.31 & 27.21 & 54.24 & 76.11 \\
\hspace{0.6em}+\,OPD & 56.46 & 80.00 & 50.83 & 63.33 & 92.81 & 95.00 & 27.90 & 33.09 & 99.04 & 98.92 \\
\hspace*{0.6em}+\,ExOPD & 57.19 & 83.33 & 52.50 & 63.33 & 93.13 & 97.50 & 28.08 & \textbf{33.27} & 100.41 & 100.78 \\
\hspace*{0.6em}+\,REOPOLD & 56.98 & 83.33 & 51.67 & 63.33 & 93.59 & 97.50 & 27.80 & 32.35 & 99.79 & 100.09 \\
\hspace*{0.6em}+\,AOPD & 56.56 & \textbf{86.67} & 52.50 & 66.67 & 93.28 & 97.50 & 27.67 & 33.09 & 99.81 & 102.98 \\
\hspace*{0.6em}+\,TOPD & 54.79 & 80.00 & 51.67 & 63.33 & 93.44 & 97.50 & 26.84 & 31.99 & 97.94 & 98.73 \\
\hspace*{0.6em}+\,OWPO & 58.33 & 80.00 & 53.85 & \textbf{76.67} & \textbf{93.75} & \textbf{97.50} & 27.76 & 31.99 & 101.44 & 103.73 \\
\hspace*{0.6em}+\,$\gamma$OPD & 60.94 & \textbf{86.67} & 54.17 & 66.67 & 92.89 & 97.50 & 27.71 & 32.90 & 102.46 & 102.84 \\
\rowcolor{cyan!8}
\hspace*{0.6em}+\,\textbf{MOPA (ours)} & \textbf{61.88} & 83.33 & \textbf{54.79} & \textbf{76.67} & 93.62 & \textbf{97.50} & \textbf{28.49} & 31.99 & \textbf{104.06} & \textbf{104.82} \\
\midrule
\multicolumn{11}{l}{\textit{Student: Qwen3-1.7B}} \\
Base & 11.77 & 46.67 & 8.33 & 20.00 & 39.84 & 90.00 & 13.88 & 22.06 & 32.27 & 62.28 \\
\hspace*{0.6em}+\,OPD & 37.71 & 66.67 & 31.25 & 40.00 & 76.02 & 92.50 & 23.16 & 29.78 & 72.66 & 82.71 \\
\hspace*{0.6em}+\,ExOPD & 38.54 & 66.67 & 31.67 & 43.33 & 76.72 & 92.50 & 23.94 & 29.23 & 74.11 & 83.54 \\
\hspace*{0.6em}+\,REOPOLD & 37.50 & 66.67 & 30.83 & 43.33 & 77.73 & 92.50 & 22.15 & 29.23 & 71.91 & 83.54 \\
\hspace*{0.6em}+\,AOPD & 37.71 & 70.00 & 32.50 & 43.33 & 79.38 & 95.00 & 22.15 & 28.31 & 73.25 & 84.58 \\
\hspace*{0.6em}+\,TOPD & 39.38 & 70.00 & 30.83 & 36.67 & 79.14 & 95.00 & 23.25 & 27.94 & 74.10 & 81.81 \\
\hspace*{0.6em}+\,OWPO & 41.67 & 70.00 & 30.83 & 46.67 & 80.47 & 95.00 & 23.99 & 30.15 & 76.11 & 87.21 \\
\hspace*{0.6em}+\,$\gamma$OPD & \textbf{42.08} & 70.00 & \textbf{35.00} & 46.67 & \textbf{80.86} & \textbf{95.00} & 23.98 & 30.14 & 78.40 & 87.20 \\
\rowcolor{cyan!8}
\hspace*{0.6em}+\,\textbf{MOPA (ours)} & 41.73 & \textbf{73.33} & 34.12 & \textbf{50.00} & 80.45 & \textbf{95.00} & \textbf{24.99} & \textbf{31.51} & \textbf{78.62} & \textbf{90.56} \\
\bottomrule
\end{tabular}
\endgroup
\end{table*}

To instantiate MOPA, we specify a temperature-controlled reliability mapping
that converts signal magnitudes into directional evidence. A shared
temperature $\tau>0$ controls the strength of evidence, while positive
source weights $w_j$ adjust the relative scales of the three signals.
Specifically, for $j\in\{\mathrm V,\mathrm O,\mathrm D\}$, we define
\begin{equation}
p_j(R_j)=\frac{\tau+w_j|R_j|}{2\tau+w_j|R_j|}.
\label{eq:reliability-map}
\end{equation}
For fixed nonzero signals, reducing $\tau$ toward zero drives reliability
toward one; relative magnitude differences then have less influence on the
aggregated direction. The weight $w_j$ adjusts the relative contribution
of source $j$. Reliability increases from
$1/2$ toward one with the weighted magnitude $w_j|R_j|/\tau$ and is strictly
greater than $1/2$ for every nonzero signal.
The corresponding evidence is
\begin{equation}
e_j(R_j)=\log\left(1+\frac{w_j|R_j|}{\tau}\right).
\label{eq:evidence-map}
\end{equation}

Substituting this evidence into the posterior mean introduced in
Section~\ref{sec:fusion-theory} gives the MOPA surrogate advantage directly
in terms of the observed signals:
\begin{equation}
\widehat A=\tanh\!\left(
\frac12\sum_{j}
\operatorname{sgn}(R_j)
\log\!\left(1+\frac{w_j|R_j|}{\tau}\right)
\right).
\label{eq:beta-advantage}
\end{equation}
The temperature parameter \(\tau\) interpolates between the two limiting behaviors:
\begin{equation}
\lim_{\tau\to\infty}2\tau\widehat A_\tau
=\sum_j w_jR_j,
\label{eq:high-temperature-limit}
\end{equation}
\begin{equation}
\lim_{\tau\to0^+}\widehat A_\tau
=\operatorname{sgn}(\sum_j\operatorname{sgn}(R_j)).
\label{eq:low-temperature-limit}
\end{equation}
At high temperatures, the surrogate advantage is
asymptotically proportional to the weighted signal sum. At low temperatures, for
three nonzero signals, the surrogate approaches the majority sign with
unit magnitude. Relative magnitude differences become less influential
as the temperature decreases toward zero.
Figure~\ref{fig:bids-temperature} illustrates this transition across relative
signal magnitudes with fixed conflicting signs. Formal limits and derivations appear in
Appendix~\ref{app:fusion-properties}.

\section{Experiments}
\label{sec:experiments}

\subsection{Experimental Settings}
\label{sec:experimental-settings}

\begin{table*}[t] \centering
\caption{\textbf{Multi-teacher distillation from Qwen3-4B-Math and Qwen3-4B-Code into Qwen3-4B.} The average scores of individual benchmarks ($\dagger$) are normalized by the corresponding teacher model's performance on each task before averaging.}
\label{tab:bids-multi}
\begingroup
\setlength{\tabcolsep}{0pt}
\newcolumntype{I}{!{\color{black!20}\vrule width 0.4pt}}
\newcolumntype{C}[1]{>{\centering\arraybackslash}m{#1}}
\begin{tabular}{l<{\hspace{15pt}}I C{0.10\textwidth} C{0.10\textwidth} C{0.10\textwidth} C{0.10\textwidth}I C{0.10\textwidth} C{0.10\textwidth} C{0.10\textwidth}I C{0.10\textwidth}}
\toprule
\multirow{2}{*}{Model} & \multicolumn{4}{cI}{Mathematical reasoning} & \multicolumn{3}{cI}{Code reasoning} & \multirow{2}{*}{$\text{TotalAvg}^\dagger$} \\
\cmidrule(lr){2-5}\cmidrule(lr){6-8}
& AIME24 & AIME25 & AMC23 & Minerva. & HumanEval+ & MBPP+ & LCB & \\
\midrule
Teacher & 57.40 & 51.67 & 93.52 & 27.85 & 82.32 & 70.63 & 25.71 & 100.00 \\
Student & 22.60 & 20.83 & 60.16 & 20.31 & 75.61 & 61.90 & 17.14 & 68.14 \\
\hspace*{0.6em}+\,OPD & 55.21 & 52.50 & 93.13 & 27.57 & 84.15 & 67.99 & 26.00 & 99.48 \\
\hspace*{0.6em}+\,ExOPD & 57.50 & 52.50 & 92.34 & 27.90 & 85.37 & 71.16 & 26.71 & 101.48 \\
\hspace*{0.6em}+\,REOPOLD & 58.13 & 52.08 & 91.88 & 27.71 & 82.32 & 69.31 & 28.00 & 101.15 \\
\hspace*{0.6em}+\,AOPD & 55.52 & \textbf{55.83} & 93.05 & 27.30 & 84.76 & 69.05 & 26.71 & 101.06 \\
\hspace*{0.6em}+\,TOPD & 57.81 & 51.67 & 92.34 & 27.02 & 84.15 & 69.58 & 27.86 & 101.08 \\
\hspace*{0.6em}+\,OWPO & 57.14 & 52.14 & 93.55 & 27.48 & 87.20 & 67.72 & 28.14 & 101.77 \\
\hspace*{0.6em}+\,$\gamma$OPD & 59.27 & 55.00 & \textbf{94.38} & 28.12 & 89.02 & 70.63 & 28.14 & 104.38 \\
\rowcolor{cyan!8}
\hspace*{0.6em}+\,\textbf{MOPA(ours)} & \textbf{59.38} & 54.37 & 94.12 & \textbf{28.95} & \textbf{89.41} & \textbf{72.49} & \textbf{29.14} & \textbf{105.76} \\
\bottomrule
\end{tabular}
\endgroup
\end{table*}

\paragraph{Baselines.}
We compare our proposed MOPA against vanilla OPD~\citep{lu2025opd},
ExOPD~\citep{yang2026gkd}, REOPOLD~\citep{ko2026reopold},
AOPD~\citep{jia2026aopd}, TOPD~\citep{zhang2026topd},
OWPO~\citep{yang2026owpo}, and
$\gamma$OPD~\citep{liu2026gammaopd}.
Appendix~\ref{app:rlvr-results} additionally compares MOPA with
the verifier-only RLVR methods JustRL~\citep{he2026justrl} and STAPO~\citep{liu2026stapo}.
For MOPA, we use the default hyperparameters $\tau=4^{-1}$, $w_V=1$, $w_O=8^{-1}$, $w_D=32^{-1}$.
All methods are implemented based on the \texttt{veRL} framework~\citep{sheng2025hybridflow}.

\paragraph{Models.}
For \emph{single-domain mathematical reasoning}, we use
Qwen3-4B-Math~\citep{yang2026gkd} as the teacher and Qwen3-4B or
Qwen3-1.7B~\citep{yangQwen3TechnicalReport2025} as the student,
corresponding to same-size and size-mismatched distillation, respectively.
For \emph{multi-domain mathematical and code reasoning}, Qwen3-4B-Math
and Qwen3-4B-Code jointly teach a single Qwen3-4B student.
Both domain teachers are obtained by applying domain-specific RL to
Qwen3-4B in non-thinking mode~\citep{yang2026gkd}; their shared teacher
base is the Qwen3-4B checkpoint before this RL training.

\paragraph{Training.}
For mathematics, we train on DeepMath-103K problems with difficulty
scores of at least 6~\citep{he2025deepmath}, using a DAPO-style
boxed-answer verifier~\citep{yu2025dapo}.
For code, we train on the 25K-example Eurus-RL-Code
dataset~\citep{cui2026prime}, using execution-based unit tests for verification.
We sample one response per prompt and compute all three training signals.
Further training details appear in Appendix~\ref{app:experiment-details}.

\paragraph{Evaluation.}
We evaluate mathematics on AIME24 and AMC23~\citep{li2024numinamath},
AIME25~\citep{opencompass2025aime}, and MinervaMath~\citep{lewkowycz2022solving}.
We sample 32 responses per problem for the competition benchmarks and
four for MinervaMath. For code, we use HumanEval+ and MBPP+~\citep{liu2023evalplus}
with one response per problem and LiveCodeBench v6~\citep{jain2025livecodebench}
with four, covering problems released from February to May 2025.
Sampling temperature and top-p are both 1.0.
Further evaluation details appear in Appendix~\ref{app:experiment-details}.

\subsection{Reasoning Performance}
\label{sec:reasoning-performance}

\begin{figure*}[t]
\centering
\includegraphics[width=0.96\linewidth]{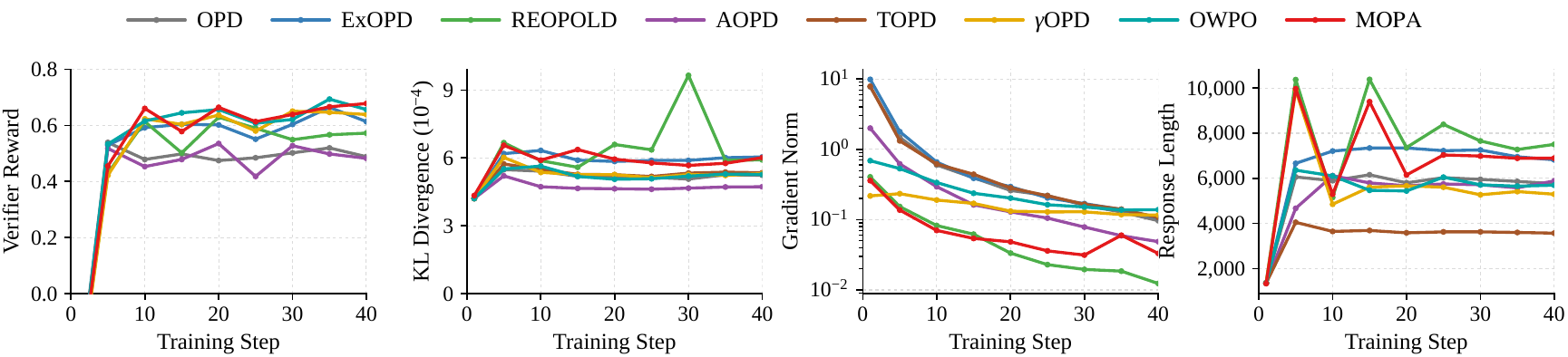}
\caption{\textbf{Optimization behavior under same-size distillation.} From left to right: verifier reward, KL divergence, gradient norm, and response length in tokens.}
\label{fig:bids-dynamics}
\end{figure*}

\paragraph{Same-size distillation.}
With Qwen3-4B-Math teaching Qwen3-4B, MOPA reaches average accuracy and
pass-rate scores of 104.06 and 104.82, respectively
(Table~\ref{tab:bids-math}). These exceed the strongest baseline for each
metric by 1.60 points over $\gamma$OPD and 1.09 points over OWPO.
Across all four benchmarks, MOPA achieves the best or joint-best result
in at least one of accuracy and pass rate among the compared distillation
methods. It leads in accuracy on AIME24, AIME25, and MinervaMath, and
ties for the best pass rate on AIME25 and AMC23. MOPA obtains these gains
through a common probabilistic mapping and fusion rule across all three
signals. Its aggregate advantage over specialized distillation methods
such as AOPD and $\gamma$OPD highlights the value of effectively
integrating additional complementary supervision.

\paragraph{Size-mismatched distillation.}
With the same teacher and a smaller Qwen3-1.7B student, MOPA reaches
average accuracy and pass-rate scores of 78.62 and 90.56
(Table~\ref{tab:bids-math}), exceeding the strongest baselines by
0.22 points over $\gamma$OPD and 3.35 points over OWPO, respectively.
MOPA achieves the best pass rate on AIME24, AIME25, and MinervaMath,
and ties for the best on AMC23 among the compared distillation methods.
On the challenging AIME24 and AIME25 benchmarks, its pass rates reach
73.33\% and 50.00\%, respectively, outperforming every compared
distillation baseline. Together with its highest accuracy on MinervaMath,
these results demonstrate the benefit of multi-signal fusion for solving
challenging reasoning problems with a substantially smaller student.

\paragraph{Multi-teacher distillation.}
When Qwen3-4B learns jointly from mathematical and code teachers, MOPA
achieves a TotalAvg of 105.76, exceeding the strongest baseline,
$\gamma$OPD, by 1.38 points (Table~\ref{tab:bids-multi}). Relative to this
baseline, the normalized domain averages improve by 0.42 points in
mathematics and 2.33 points in code. MOPA achieves the highest scores
on five of the seven benchmarks, including all three code benchmarks
as well as AIME24 and MinervaMath. These results demonstrate effective
learning from domain-specialized teachers, with aggregate gains in
both mathematical and code reasoning.

\subsection{Training Dynamics}
\label{sec:training-dynamics}

\paragraph{Optimization behavior.}
Figure~\ref{fig:bids-dynamics} shows that MOPA reaches a positive verifier
reward within the first five steps and maintains a strong trajectory after
the initial transition. Its mean reward over the sampled checkpoints from
step 30 to step 40 is 0.661, compared with 0.645 for $\gamma$OPD, and it
finishes at 0.678. This reward improvement does not coincide with unstable
optimization. The KL divergence settles near $6\times10^{-4}$, within the
range of the comparison methods, while the gradient norm decreases from
0.35 at the first step to 0.033 at step 40. Its late-stage gradient norm is
0.041 on average, which is 65.8\% lower than that of $\gamma$OPD over the
same checkpoints. Response length fluctuates during the first 15 steps and
then stabilizes near 6.9K tokens. This length is close to ExOPD and below
REOPOLD, although it remains about 30\% above $\gamma$OPD. Overall, MOPA
combines the strongest late-stage verifier reward with bounded policy drift
and a steadily decreasing update magnitude.

\begin{figure}[h]
\centering
\includegraphics[width=0.92\linewidth]{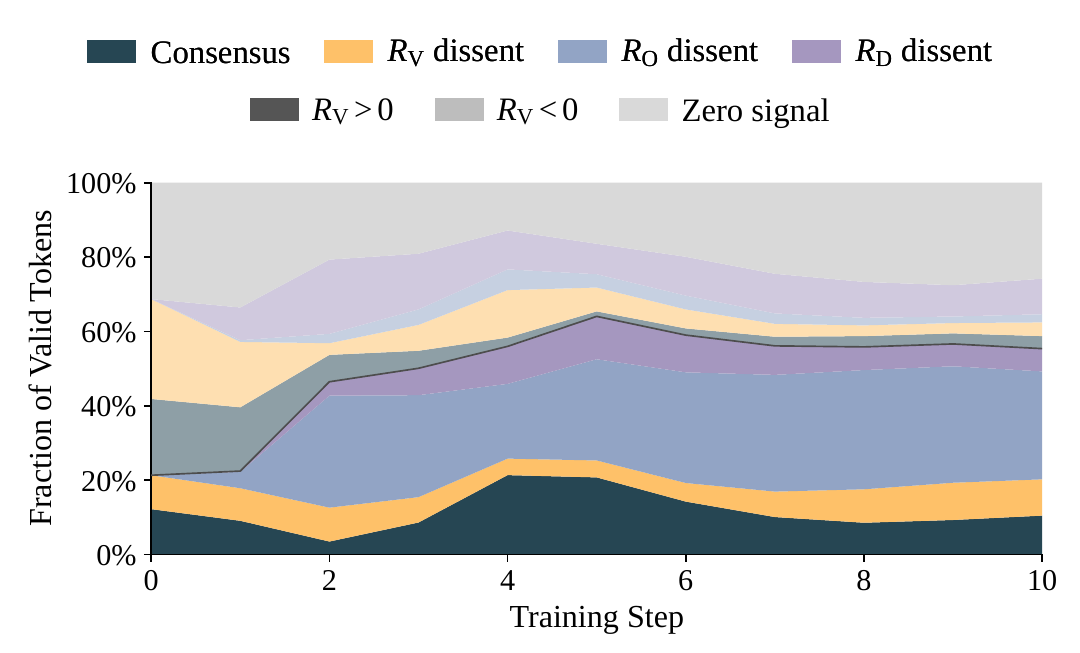}
\caption{\textbf{Evolution of signal directions during early MOPA training.}
Stacked areas show fractions of valid tokens over steps 0–10. The eight colored regions represent the sign combinations of the three signals, grouped by positive verifier outcomes below and negative outcomes above. Colors distinguish consensus from dissent by each source. The gray region indicates tokens with at least one zero-valued signal.}
\label{fig:bids-octant-ratios}
\end{figure}

\paragraph{Directional composition.}
Figure~\ref{fig:bids-octant-ratios} shows the early evolution of the
directional composition, when its changes are most pronounced. After the
first ten steps, the composition remains nearly unchanged. In the resulting near-steady regime, only 20.1\% of tokens with three nonzero signals exhibit consensus.
The remaining 79.9\% contain a dissenting source, so direction fusion governs
most token updates. Negative consensus decreases particularly quickly during
the early phase. Tokens opposed by all three signals fall
from 20.5\% at the first step to approximately 2\% within five steps and
remain rare thereafter. This pattern is consistent with rapid suppression
of unambiguously unfavorable tokens. Later negative responses are therefore
dominated by conflicting evidence.
The most common conflict comes from the teacher--student discrepancy signal
and accounts for 41.8\% of tokens with three nonzero signals. Among these
cases, 90.6\% occur in verifier-positive responses. The sampled token has a
higher likelihood under the student than under the teacher in this regime,
while both the verifier and the teacher-adaptation signal support
its direction. This pattern indicates that most observed extrapolation is
beneficial. MOPA preserves these updates and avoids an automatic pull back
toward the teacher.

\begin{figure*}[t]
\centering
\vspace{-5pt}
\subfigure[Signal sources.]{
  \includegraphics[width=0.46\linewidth]{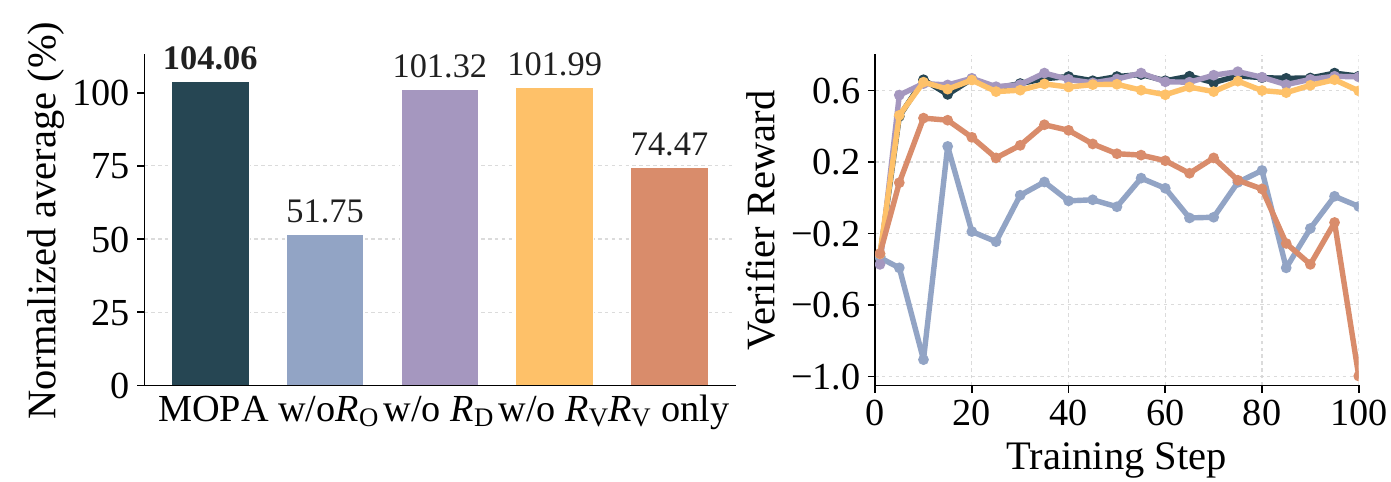}
  \label{fig:bids-ablation-signals}}
\hfill
\vspace{-5pt}
\subfigure[Temperature.]{
  \includegraphics[width=0.245\linewidth]{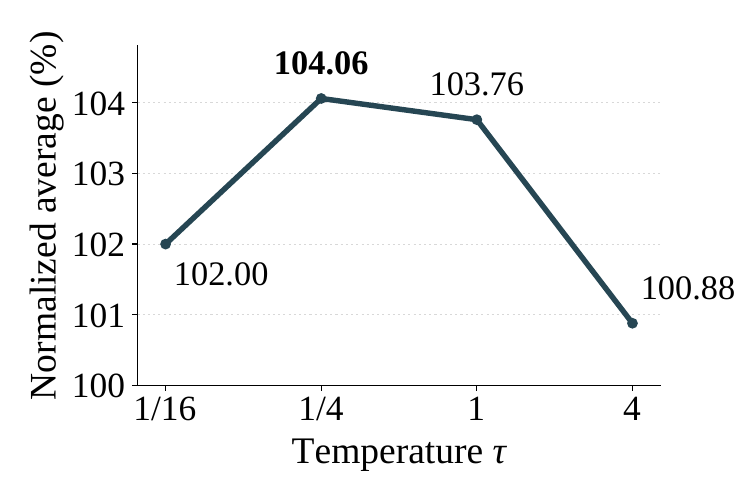}
  \label{fig:bids-ablation-temperature}}
\hspace{-5pt}
\hfill
\vspace{-5pt}
\subfigure[Relative source weight.]{
  \includegraphics[width=0.245\linewidth]{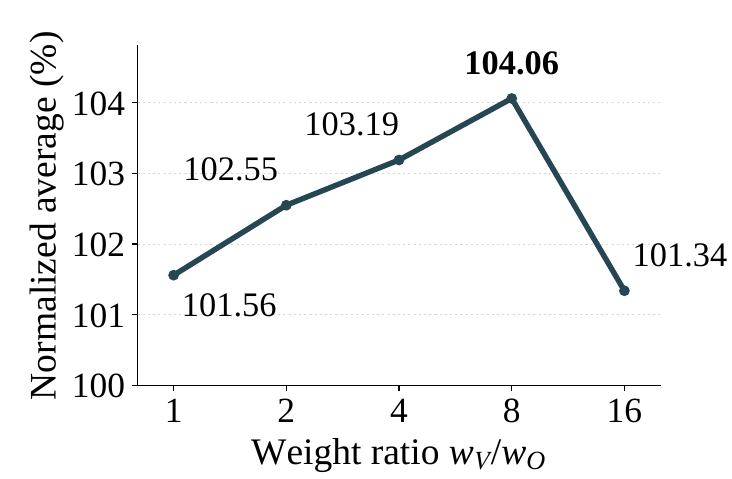}
  \label{fig:bids-ablation-weights}}
\caption{\textbf{MOPA ablations under same-size distillation.}
(a) Accuracy and verifier reward for different signal subsets. (b) Effect of temperature \(\tau\) with fixed source-weight ratios. (c) Effect of \(w_{\mathrm V}/w_{\mathrm O}\) with \(w_{\mathrm O}/w_{\mathrm D}=4\). Accuracy is averaged over the four mathematical benchmarks after teacher normalization, with teacher performance set to 100.}
\label{fig:bids-ablation}
\end{figure*}

\subsection{Ablation Study}
\label{sec:bids-ablation}

We ablate the signal sources, evidence temperature, and relative source
weights under the vanilla-distillation setting. Figure~\ref{fig:bids-ablation}
reports teacher-normalized average accuracy as in Table~\ref{tab:bids-math},
using the best completed checkpoint evaluation for each setting.

\paragraph{Complementarity of the three signals.}
Figure~\ref{fig:bids-ablation-signals} compares the accuracy and training
reward of signal subsets under fixed fusion hyperparameters, yielding
three observations about their complementary roles.
\textit{(1) Teacher signals remain useful even when the verifier sets the direction.}
Under the best-performing configuration, the fused update direction always
agrees with $R_{\mathrm V}$ throughout training. Nevertheless,
verifier-only training collapses while the full combination sustains
learning. By Equation~\eqref{eq:beta-advantage}, the teacher signals
$R_{\mathrm O}$ and $R_{\mathrm D}$ still modulate token-level update
magnitudes, strengthening updates where their evidence agrees with the
verifier and attenuating them where it conflicts. They thus supplement
response-level direction with local variation in update strength.
\textit{(2) The teacher--student discrepancy signal contributes the most despite its lower weight.}
Although $R_{\mathrm O}$ receives only one-eighth of the verifier's weight
($w_{\mathrm O}=1/8$ versus $w_{\mathrm V}=1$), removing it from the full
combination reduces teacher-normalized average accuracy by 52.31 points.
In contrast, removing either $R_{\mathrm V}$ or $R_{\mathrm D}$ reduces
accuracy by fewer than three points. Thus, a source's fusion weight does
not directly reflect its contribution to learning: the teacher--student
discrepancy signal is the most consequential in these ablations, while
verifier feedback and teacher adaptation provide smaller, complementary gains.
\textit{(3) Teacher adaptation helps when combined with the teacher--student discrepancy signal.}
Adding $R_{\mathrm D}$ to $R_{\mathrm V}$ alone reduces performance,
whereas adding it to $R_{\mathrm V}+R_{\mathrm O}$ yields a gain.
The benefit of teacher adaptation therefore depends on the accompanying
teacher--student discrepancy signal in these experiments.

\paragraph{Temperature robustness and effects.}
With source-weight ratios fixed, Figure~\ref{fig:bids-ablation-temperature}
shows that MOPA outperforms the strongest baseline across a broad range of
tested temperatures, with similar performance at $\tau=1/4$ and $\tau=1$,
indicating limited sensitivity to temperature tuning. Performance declines
at $\tau=1/16$ and $\tau=4$, which approximate the majority-voting and
linear-aggregation regimes, respectively. As discussed in
Section~\ref{sec:evidence-fusion}, a small temperature suppresses relative
magnitude information, allowing weak signals to cast nearly full votes;
a large temperature weakens logarithmic evidence compression, making the
fused direction more susceptible to large-magnitude signals. Intermediate
temperatures retain useful magnitude information while tempering the
influence of dominant signals, enabling more selective resolution of
directional conflicts.

\paragraph{Relative source weighting.}
Following the best ExOPD setting, we fix $w_{\mathrm O}/w_{\mathrm D}=4$
and vary the verifier-to-OPD weight ratio. Figure~\ref{fig:bids-ablation-weights}
shows that source balance, rather than simply increasing verifier influence,
is important. Performance improves as this ratio grows from 1 to 8, but
declines at 16. Together with the source ablation, this non-monotonic trend
indicates that response-level correctness should steer the dense teacher-student 
signal without overriding it.

\section{Conclusion}

We introduced MOPA for combining conflicting signals in LLM
post-training.
MOPA models verifier outcomes, teacher--student discrepancies, and
teacher-adaptation signals as evidence about a shared advantage direction,
then uses the posterior mean direction as a bounded surrogate advantage.
Its temperature-controlled logarithmic mapping adjusts the influence of
directional agreement relative to signal magnitude, connecting majority
voting and scaled linear aggregation under the stated conditions. 
Experiments on mathematical and code reasoning show improved
teacher-normalized aggregate performance across the evaluated
same-size, size-mismatched, and multi-teacher settings.
The training analysis further shows stable policy improvement
and decreasing gradient norms, while the ablations confirm that dense teacher-student
guidance is the primary anchor and that verifier and teacher-adaptation signals
provide complementary gains when their relative influence is balanced.

MOPA assumes conditionally independent signal directions and uses globally
tuned source weights and temperature, which may not capture correlated or
state-dependent signal reliability. Future work will evaluate MOPA with a broader range of supervision signals and investigate adaptive fusion that accounts for signal reliability and dependence.

\nocite{aid}
\nocite{lyu2025conformal}
\nocite{zheng2026past}

\bibliography{example_paper}

\begin{thebibliography}{41}
\providecommand{\natexlab}[1]{#1}
\providecommand{\url}[1]{\texttt{#1}}
\expandafter\ifx\csname urlstyle\endcsname\relax
  \providecommand{\doi}[1]{doi: #1}\else
  \providecommand{\doi}{doi: \begingroup \urlstyle{rm}\Url}\fi

\bibitem[Agarwal et~al.(2024)Agarwal, Vieillard, Zhou, Stanczyk, Ramos, Geist, and Bachem]{agarwal2024gkd}
Agarwal, R., Vieillard, N., Zhou, Y., Stanczyk, P., Ramos, S., Geist, M., and Bachem, O.
\newblock On-policy distillation of language models: Learning from self-generated mistakes.
\newblock In \emph{International Conference on Learning Representations}, 2024.

\bibitem[Akhondzadeh et~al.(2026)Akhondzadeh, Lingam, Tejaswi, Ekbote, Sanghavi, and Bojchevski]{akhondzadeh2026rgopd}
Akhondzadeh, M.~S., Lingam, V., Tejaswi, A., Ekbote, C., Sanghavi, S., and Bojchevski, A.
\newblock Reward-gated on-policy distillation.
\newblock \emph{arXiv preprint arXiv:2607.04037}, 2026.

\bibitem[Cui et~al.(2026)Cui, Yuan, Wang, Wang, Zhang, Chen, et~al.]{cui2026prime}
Cui, G., Yuan, L., Wang, Z., Wang, H., Zhang, Y., Chen, J., et~al.
\newblock Process reinforcement through implicit rewards.
\newblock \emph{Transactions on Machine Learning Research}, 2026.

\bibitem[Ding et~al.(2026)Ding, Wei, Li, Li, Lu, et~al.]{ding2026saf}
Ding, Y., Wei, X., Li, Y.~Y., Li, Z., Lu, Y., et~al.
\newblock Saf-opd: Stable advantage fusion for on-policy distillation.
\newblock \emph{arXiv preprint arXiv:2607.29209}, 2026.

\bibitem[Feng et~al.(2026)Feng, Gao, Chi, Wu, Zhang, et~al.]{feng2026directopd}
Feng, S., Gao, H.-a., Chi, H., Wu, H., Zhang, Z., et~al.
\newblock Weak-to-strong generalization via direct on-policy distillation.
\newblock \emph{arXiv preprint arXiv:2607.05394}, 2026.

\bibitem[Fu et~al.(2026)Fu, Huang, Jiang, Liu, Jiang, Zhu, and Zhao]{fu2026revisiting}
Fu, Y., Huang, H., Jiang, K., Liu, J., Jiang, Z., Zhu, Y., and Zhao, D.
\newblock Revisiting on-policy distillation: Empirical failure modes and simple fixes.
\newblock \emph{Conference on Language Modeling}, 2026.

\bibitem[Guan et~al.(2021)Guan, Li, Duan, Li, Ren, Sun, and Cheng]{guan2021direct}
Guan, Y., Li, S.~E., Duan, J., Li, J., Ren, Y., Sun, Q., and Cheng, B.
\newblock Direct and indirect reinforcement learning.
\newblock \emph{International Journal of Intelligent Systems}, 2021.

\bibitem[Guo et~al.(2025)Guo, Yang, Zhang, Song, Zhang, Xu, et~al.]{deepseekr12025}
Guo, D., Yang, D., Zhang, H., Song, J., Zhang, R., Xu, R., et~al.
\newblock Deepseek-r1: Incentivizing reasoning capability in llms via reinforcement learning.
\newblock \emph{arXiv preprint arXiv:2501.12948}, 2025.

\bibitem[He et~al.(2026)He, Qu, Liu, Chen, Zuo, Qian, Zhang, Chen, Xiao, Cui, Ding, and Liu]{he2026justrl}
He, B., Qu, Z., Liu, Z., Chen, Y., Zuo, Y., Qian, C., Zhang, K., Chen, W., Xiao, C., Cui, G., Ding, N., and Liu, Z.
\newblock {JustRL}: Scaling a 1.5b {LLM} with a simple {RL} recipe.
\newblock In \emph{ICLR Blogposts 2026}, 2026.

\bibitem[He et~al.(2025)He, Liang, Xu, Liu, Chen, Wang, Song, Yu, Liang, Wang, Zhang, Wang, Tu, Mi, and Yu]{he2025deepmath}
He, Z., Liang, T., Xu, J., Liu, Q., Chen, X., Wang, Y., Song, L., Yu, D., Liang, Z., Wang, W., Zhang, Z., Wang, R., Tu, Z., Mi, H., and Yu, D.
\newblock {DeepMath-103K}: A large-scale, challenging, decontaminated, and verifiable mathematical dataset for advancing reasoning.
\newblock \emph{arXiv preprint arXiv:2504.11456}, 2025.

\bibitem[Heo et~al.(2026)Heo, Hwang, Yun, and Han]{heo2026opd2}
Heo, B., Hwang, J., Yun, S., and Han, D.
\newblock On-policy delta distillation.
\newblock \emph{arXiv preprint arXiv:2607.15161}, 2026.

\bibitem[Hou et~al.(2026)Hou, Peng, Wang, Ruan, Zhang, et~al.]{hou2026uniopd}
Hou, W., Peng, S., Wang, W., Ruan, Z., Zhang, Y., et~al.
\newblock Uni-opd: Unifying on-policy distillation with a dual-perspective recipe.
\newblock \emph{arXiv preprint arXiv:2605.03677}, 2026.

\bibitem[Jain et~al.(2025)Jain, Han, Gu, Li, Yan, Zhang, Wang, Solar-Lezama, Sen, and Stoica]{jain2025livecodebench}
Jain, N., Han, K., Gu, A., Li, W.-D., Yan, F., Zhang, T., Wang, S., Solar-Lezama, A., Sen, K., and Stoica, I.
\newblock {LiveCodeBench}: Holistic and contamination free evaluation of large language models for code.
\newblock In \emph{International Conference on Learning Representations}, 2025.

\bibitem[Jia et~al.(2026)Jia, Yang, Ma, Lian, Zhang, Zhang, Zeng, Cai, and Sun]{jia2026aopd}
Jia, N., Yang, H., Ma, X., Lian, J., Zhang, S., Zhang, W., Zeng, K., Cai, X., and Sun, Z.
\newblock Asymmetric on-policy distillation: Bridging exploitation and imitation at the token level.
\newblock \emph{arXiv preprint arXiv:2605.06387}, 2026.

\bibitem[Ko et~al.(2026)Ko, Abdali, Kim, Chen, and Cameron]{ko2026reopold}
Ko, J., Abdali, S., Kim, Y.~J., Chen, T., and Cameron, P.
\newblock Relaxed on-policy distillation: Selective credit allocation for scaling reasoning efficiently.
\newblock In \emph{International Conference on Machine Learning}, 2026.

\bibitem[Lewkowycz et~al.(2022)Lewkowycz, Andreassen, Dohan, Dyer, Michalewski, Ramasesh, Slone, Anil, Schlag, Gutman-Solo, et~al.]{lewkowycz2022solving}
Lewkowycz, A., Andreassen, A., Dohan, D., Dyer, E., Michalewski, H., Ramasesh, V., Slone, A., Anil, C., Schlag, I., Gutman-Solo, T., et~al.
\newblock Solving quantitative reasoning problems with language models.
\newblock \emph{Advances in neural information processing systems}, 2022.

\bibitem[Li et~al.(2024)Li, Beeching, Tunstall, Lipkin, Soletskyi, Huang, Rasul, Yu, Jiang, Shen, et~al.]{li2024numinamath}
Li, J., Beeching, E., Tunstall, L., Lipkin, B., Soletskyi, R., Huang, S., Rasul, K., Yu, L., Jiang, A.~Q., Shen, Z., et~al.
\newblock {NuminaMath}: The largest public dataset in {AI4Maths} with 860k pairs of competition math problems and solutions.
\newblock Hugging Face dataset, 2024.
\newblock URL \url{https://huggingface.co/datasets/AI-MO/NuminaMath-CoT}.

\bibitem[Li(2023)]{li2023reinforcement}
Li, S.~E.
\newblock \emph{Reinforcement learning for sequential decision and optimal control}.
\newblock Springer, 2023.

\bibitem[Li et~al.(2026)Li, Zuo, He, Zhang, Xiao, et~al.]{li2026rethinking}
Li, Y., Zuo, Y., He, B., Zhang, J., Xiao, C., et~al.
\newblock Rethinking on-policy distillation of large language models: Phenomenology, mechanism, and recipe.
\newblock \emph{arXiv preprint arXiv:2604.13016}, 2026.

\bibitem[Lin et~al.(2026)Lin, Zhao, Jiang, Rao, Li, Wang, He, and Huang]{lin2026opdvr}
Lin, W., Zhao, J., Jiang, X., Rao, S., Li, Y., Wang, S., He, B., and Huang, G.
\newblock On-policy distillation with verifiable reward.
\newblock \emph{arXiv preprint arXiv:2608.24696}, 2026.

\bibitem[Liu et~al.(2023)Liu, Xia, Wang, and Zhang]{liu2023evalplus}
Liu, J., Xia, C.~S., Wang, Y., and Zhang, L.
\newblock Is your code generated by {ChatGPT} really correct? rigorous evaluation of large language models for code generation.
\newblock In \emph{Advances in Neural Information Processing Systems}, 2023.

\bibitem[Liu et~al.(2026{\natexlab{a}})Liu, He, Tao, Zhan, Gao, Zhang, Duan, Xiong, Sheng, Zhang, et~al.]{liu2026gammaopd}
Liu, S., He, Z., Tao, L., Zhan, G., Gao, J., Zhang, F., Duan, J., Xiong, W., Sheng, K., Zhang, B., et~al.
\newblock Beyond token-local imitation: Reward-compatible temporal credit assignment for on-policy distillation.
\newblock \emph{arXiv preprint arXiv:2609.16937}, 2026{\natexlab{a}}.

\bibitem[Liu et~al.(2026{\natexlab{b}})Liu, He, Zhan, Tao, Zheng, et~al.]{liu2026stapo}
Liu, S., He, Z., Zhan, G., Tao, L., Zheng, Z., et~al.
\newblock Stapo: Stabilizing reinforcement learning for llms by silencing rare spurious tokens.
\newblock \emph{Advances in Neural Information Processing Systems}, 2026{\natexlab{b}}.

\bibitem[Lu \& {Thinking Machines Lab}(2025)Lu and {Thinking Machines Lab}]{lu2025opd}
Lu, K. and {Thinking Machines Lab}.
\newblock On-policy distillation.
\newblock Thinking Machines Lab: Connectionism, 2025.
\newblock URL \url{https://thinkingmachines.ai/blog/on-policy-distillation}.

\bibitem[Lyu et~al.(2025)Lyu, Zhang, Li, Duan, Tao, Xu, He, and Li]{lyu2025conformal}
Lyu, Y., Zhang, X., Li, S.~E., Duan, J., Tao, L., Xu, Q., He, L., and Li, K.
\newblock Conformal symplectic optimization for stable reinforcement learning.
\newblock \emph{IEEE Transactions on Neural Networks and Learning Systems}, 2025.

\bibitem[{OpenCompass}(2025)]{opencompass2025aime}
{OpenCompass}.
\newblock {AIME2025} dataset.
\newblock Hugging Face dataset, 2025.
\newblock URL \url{https://huggingface.co/datasets/opencompass/AIME2025}.

\bibitem[Shao et~al.(2024)Shao, Wang, Zhu, Xu, Song, Zhang, et~al.]{deepseekmath2024}
Shao, Z., Wang, P., Zhu, Q., Xu, R., Song, J., Zhang, M., et~al.
\newblock Deepseekmath: Pushing the limits of mathematical reasoning in open language models.
\newblock \emph{arXiv preprint arXiv:2402.03300}, 2024.

\bibitem[Sheng et~al.(2025)Sheng, Zhang, Ye, Wu, Zhang, Zhang, Peng, Lin, and Wu]{sheng2025hybridflow}
Sheng, G., Zhang, C., Ye, Z., Wu, X., Zhang, W., Zhang, R., Peng, Y., Lin, H., and Wu, C.
\newblock {HybridFlow}: A flexible and efficient {RLHF} framework.
\newblock In \emph{Proceedings of the Twentieth European Conference on Computer Systems}, 2025.

\bibitem[Sun et~al.(2026)Sun, Ma, Qin, Liu, Lu, Zhu, Cai, and Yan]{sun2026reopd}
Sun, Y., Ma, L., Qin, H., Liu, Y., Lu, H., Zhu, Y., Cai, P., and Yan, G.
\newblock Reopd: Reliability-adaptive reward extrapolation for on-policy distillation.
\newblock \emph{arXiv preprint arXiv:2608.11698}, 2026.

\bibitem[Wang et~al.(2025)Wang, Yu, Gao, Zheng, Liu, et~al.]{wang2025entropy}
Wang, S., Yu, L., Gao, C., Zheng, C., Liu, S., et~al.
\newblock Beyond the 80/20 rule: High-entropy minority tokens drive effective reinforcement learning for llm reasoning.
\newblock \emph{Advances in Neural Information Processing Systems}, 2025.

\bibitem[{Xiaomi LLM-Core Team} et~al.(2026){Xiaomi LLM-Core Team}, Xiao, Xia, Yang, Gao, et~al.]{teamMiMoV2FlashTechnicalReport2026}
{Xiaomi LLM-Core Team}, Xiao, B., Xia, B., Yang, B., Gao, B., et~al.
\newblock Mimo-v2-flash technical report.
\newblock \emph{arXiv preprint arXiv:2601.02780}, 2026.

\bibitem[Xu et~al.(2026)Xu, Wang, Gao, Li, Zhang, and Yuan]{xu2026sgopd}
Xu, H., Wang, H., Gao, Y., Li, J., Zhang, X., and Yuan, X.
\newblock Sg-opd: Sign-gated on-policy distillation via sign-consistency gating and phased teacher sampling.
\newblock \emph{arXiv preprint arXiv:2606.09304}, 2026.

\bibitem[Yang et~al.(2025)Yang, Li, Yang, Zhang, Hui, et~al.]{yangQwen3TechnicalReport2025}
Yang, A., Li, A., Yang, B., Zhang, B., Hui, B., et~al.
\newblock Qwen3 technical report.
\newblock \emph{arXiv preprint arXiv:2505.09388}, 2025.

\bibitem[Yang et~al.(2026{\natexlab{a}})Yang, Qin, Si, Chen, Gu, et~al.]{yang2026rlsd}
Yang, C., Qin, C., Si, Q., Chen, M., Gu, N., et~al.
\newblock Self-distilled rlvr.
\newblock \emph{arXiv preprint arXiv:2604.03128}, 2026{\natexlab{a}}.

\bibitem[Yang et~al.(2026{\natexlab{b}})Yang, Lu, Huang, Ma, Wei, et~al.]{yang2026owpo}
Yang, S., Lu, J., Huang, K., Ma, C., Wei, S., et~al.
\newblock One-way policy optimization for self-evolving llms.
\newblock In \emph{International Conference on Machine Learning}, 2026{\natexlab{b}}.

\bibitem[Yang et~al.(2026{\natexlab{c}})Yang, Liu, Xie, Yang, Yang, and Lin]{yang2026gkd}
Yang, W., Liu, W., Xie, R., Yang, K., Yang, S., and Lin, Y.
\newblock Learning beyond teacher: Generalized on-policy distillation with reward extrapolation.
\newblock \emph{arXiv preprint arXiv:2602.12125}, 2026{\natexlab{c}}.

\bibitem[Yu et~al.(2025)Yu, Zhang, Zhu, Yuan, Zuo, et~al.]{yu2025dapo}
Yu, Q., Zhang, Z., Zhu, R., Yuan, Y., Zuo, X., et~al.
\newblock Dapo: An open-source llm reinforcement learning system at scale.
\newblock \emph{Advances in Neural Information Processing Systems}, 2025.

\bibitem[Yu et~al.(2026)Yu, Zhan, Guan, Duan, Tao, and Li]{aid}
Yu, Z., Zhan, G., Guan, Y., Duan, J., Tao, L., and Li, S.~E.
\newblock Taming aleatoric impulse in off-policy reinforcement learning.
\newblock In \emph{International Conference on Machine Learning}, 2026.

\bibitem[Zhang et~al.(2026{\natexlab{a}})Zhang, Chai, Fu, Tu, Wang, Lin, Yin, Zhang, Zhu, and Zhao]{zhang2026topd}
Zhang, Y., Chai, J., Fu, Y., Tu, S., Wang, X., Lin, W., Yin, G., Zhang, Q., Zhu, Y., and Zhao, D.
\newblock Are full rollouts necessary for on-policy distillation?
\newblock \emph{arXiv preprint arXiv:2605.31490}, 2026{\natexlab{a}}.

\bibitem[Zhang et~al.(2026{\natexlab{b}})Zhang, Wang, Xu, Wang, Zhou, et~al.]{zhang2026gcopd}
Zhang, Z., Wang, J., Xu, X., Wang, X., Zhou, Z., et~al.
\newblock Beyond teacher likelihood: Group-calibrated on-policy distillation for long-context reasoning.
\newblock \emph{arXiv preprint arXiv:2608.19181}, 2026{\natexlab{b}}.

\bibitem[Zheng et~al.(2026)Zheng, Tao, Guan, Yang, Xiong, Sheng, Zhang, Duan, Li, and Li]{zheng2026past}
Zheng, Z., Tao, L., Guan, Y., Yang, Y., Xiong, W., Sheng, K., Zhang, B., Duan, J., Li, K., and Li, S.~E.
\newblock Past, future, all at once: Mitigating {Stability-Plasticity} dilemma via post-hoc {JANUS} rectification.
\newblock \emph{Advances in Neural Information Processing Systems}, 2026.

\end{thebibliography}
\bibliographystyle{icml2026}

\newpage
\appendix
\onecolumn
\section{True Token Advantage and the Policy Gradient}
\label{app:policy-gradient-advantage}

All expectations in this section use the current policy $\pi_\theta$ and
the terminal verifier reward $R_{\mathrm V}$. The true token advantage is
\begin{equation}
A_{t}
=\mathbb E[R_{\mathrm V}\mid h_{t},y_{t}]
-\mathbb E[R_{\mathrm V}\mid h_{t}].
\label{eq:true-token-advantage}
\end{equation}
The first term averages over continuations after the sampled token; the
second also averages over that token. Thus $A_{t}>0$ means that the token
improves expected verifier reward relative to the policy's average action
at the same prefix.

The score $\nabla_\theta\log\pi_\theta(y_{t}\mid h_{t})$ depends only
on the prefix and sampled token. The tower property therefore allows
$R_{\mathrm V}$ in its expected gradient contribution to be replaced by
$\mathbb E[R_{\mathrm V}\mid h_{t},y_{t}]$ without changing the
expectation. Moreover,
\begin{equation}
\mathbb E_{y_{t}\sim\pi_\theta(\cdot\mid h_{t})}
\!\left[\nabla_\theta\log\pi_\theta(y_{t}\mid h_{t})\right]=0,
\end{equation}
so subtracting the prefix-dependent second term also leaves the gradient
expectation unchanged. Applying these identities to every token proves
Equation~\eqref{eq:advantage-policy-gradient}.

Conditioning the return on the prefix and token removes future-rollout
noise and cannot increase the variance of that token's uncentered gradient
contribution. The prefix baseline centers the conditional return for credit
assignment. This motivates the true advantage as a variance-reduction
target; it does not establish a minimum-variance estimator for the full
trajectory gradient. MOPA estimates its direction from the three observed
signals rather than computing these conditional expectations exactly.

\section{Derivations for General Bayesian Direction Inference}

\subsection{Proof of Proposition~\ref{prop:linear-evidence}}
\label{app:linear-evidence-proof}

As in Section~\ref{sec:fusion-theory}, all probabilities below are
conditioned on the observed signal magnitudes. Thus observing $R_j$
amounts to observing its direction at the fixed magnitude.
We first relate each source's reliability to its posterior directional
probabilities. By Equation~\eqref{eq:source-likelihood},
\begin{equation}
\begin{aligned}
P(A>0\mid R_j)
&=\begin{cases}
p_j(R_j), & R_j>0,\\
1-p_j(R_j), & R_j<0,
\end{cases}\\
P(A<0\mid R_j)
&=\begin{cases}
1-p_j(R_j), & R_j>0,\\
p_j(R_j), & R_j<0.
\end{cases}
\end{aligned}
\end{equation}
For a negative observation, agreement means $A<0$, so the posterior odds
between positive and negative advantage directions are inverted. Hence
\begin{equation}
\frac{P(A>0\mid R_j)}{P(A<0\mid R_j)}
=\left(\frac{p_j(R_j)}{1-p_j(R_j)}\right)^{\operatorname{sgn}(R_j)}.
\end{equation}
This identity allows reliability to depend on the signed observation and
does not require equal reliability for positive and negative signals.

Next, Bayes' rule relates these single-source posterior odds to a
likelihood ratio:
\begin{equation}
\begin{aligned}
\frac{P(A>0\mid R_j)}{P(A<0\mid R_j)}
&=\frac{P(R_j\mid A>0)P(A>0)/P(R_j)}
{P(R_j\mid A<0)P(A<0)/P(R_j)}\\
&=\frac{P(R_j\mid A>0)}{P(R_j\mid A<0)}
\frac{P(A>0)}{P(A<0)}\\
&=\frac{P(R_j\mid A>0)}{P(R_j\mid A<0)}.
\end{aligned}
\end{equation}
The last equality uses the balanced conditional prior in
Proposition~\ref{prop:linear-evidence}.

Finally, apply Bayes' rule to the joint observation. Conditional
independence of the signal directions, given the advantage direction
and observed magnitudes, factorizes each joint likelihood. Therefore,
\begin{equation}
\begin{aligned}
\frac{P(A>0\mid\boldsymbol R)}{P(A<0\mid\boldsymbol R)}
&=\frac{P(\boldsymbol R\mid A>0)}{P(\boldsymbol R\mid A<0)}
\frac{P(A>0)}{P(A<0)}\\
&=\frac{\prod_j P(R_j\mid A>0)}{\prod_j P(R_j\mid A<0)}\\
&=\prod_j\frac{P(A>0\mid R_j)}{P(A<0\mid R_j)}\\
&=\prod_{j\in\{\mathrm V,\mathrm O,\mathrm D\}}
\left(\frac{p_j(R_j)}{1-p_j(R_j)}\right)^{\operatorname{sgn}(R_j)}.
\end{aligned}
\label{eq:posterior-odds-product}
\end{equation}
Taking logarithms converts this product into
\begin{equation}
\begin{aligned}
L
&=\log\prod_j
\left(\frac{p_j(R_j)}{1-p_j(R_j)}\right)^{\operatorname{sgn}(R_j)}\\
&=\sum_j\operatorname{sgn}(R_j)
\log\frac{p_j(R_j)}{1-p_j(R_j)}\\
&=\sum_j
\operatorname{sgn}(R_j)
\left[\log p_j(R_j)-\log\bigl(1-p_j(R_j)\bigr)\right]\\
&=\sum_j\operatorname{sgn}(R_j)\,e_j(R_j),
\end{aligned}
\end{equation}
which proves Equation~\eqref{eq:general-evidence-sum} with the evidence
definition in Equation~\eqref{eq:source-evidence}. The balanced prior avoids
counting prior evidence repeatedly when combining single-source posteriors.
Zero signals are
omitted from the product and contribute zero to the sum.

\subsection{Posterior Expectation of the Advantage Direction}
\label{app:posterior-direction-proof}

The posterior mean direction satisfies
\begin{equation}
\begin{aligned}
\mathbb E[\operatorname{sgn}(A)\mid\boldsymbol R]
&=P(A>0\mid\boldsymbol R)-P(A<0\mid\boldsymbol R)\\
&=2P(A>0\mid\boldsymbol R)-1.
\end{aligned}
\end{equation}
The definition of $L$ gives $P(A>0\mid\boldsymbol R)=1/(1+e^{-L})$.
Substitution yields
\begin{equation}
2P(A>0\mid\boldsymbol R)-1
=\frac{1-e^{-L}}{1+e^{-L}}=\tanh(L/2),
\end{equation}
proving Equation~\eqref{eq:posterior-mean-direction}.

\section{Properties of Tempered Bayesian Direction Inference}
\label{app:fusion-properties}

For fixed $\tau$ and weights, Equation~\eqref{eq:beta-advantage} preserves
the evidence direction and bounds its magnitude by one for every $\tau$,
since $\tanh$ is odd, strictly increasing, and bounded in absolute value by one.
This controls the output amplitude, but does not prevent an extreme source
from changing the fused direction.

For both limits, fix the signals and positive weights $w_j$, and denote
the surrogate by $\widehat A_\tau$ and its posterior log odds by $L_\tau$.

\paragraph{Large-temperature limit.}
For each source,
$\lim_{\tau\to\infty}\tau\log(1+w_j|R_j|/\tau)=w_j|R_j|$.
Consequently,
$\lim_{\tau\to\infty}L_\tau=0$ and
$\lim_{\tau\to\infty}\tau L_\tau=\sum_j w_jR_j$.
The function $\tanh(x)/x$, extended continuously to equal one at zero,
therefore gives
\begin{equation}
\lim_{\tau\to\infty}2\tau\widehat A_\tau
=\lim_{\tau\to\infty}\left[
\tau L_\tau\,\frac{\tanh(L_\tau/2)}{L_\tau/2}
\right]
=\sum_j w_jR_j.
\label{eq:additive-limit}
\end{equation}

\paragraph{Small-temperature limit.}
Assume all three signals are nonzero and let
$M=\sum_j\operatorname{sgn}(R_j)$.
For each source,
$\lim_{\tau\to0^+}
\log(1+w_j|R_j|/\tau)/\log(1/\tau)=1$.
Thus
$\lim_{\tau\to0^+}L_\tau/\log(1/\tau)=M$.
With three nonzero signs, $M\in\{-3,-1,1,3\}$, so $L_\tau$ diverges with
sign $\operatorname{sgn}(M)$. Consequently,
\begin{equation}
\lim_{\tau\to0^+}\widehat A_\tau
=\lim_{\tau\to0^+}\tanh(L_\tau/2)
=\operatorname{sgn}(M).
\label{eq:majority-limit}
\end{equation}
Both unanimous and two-to-one
sign configurations approach a majority-directed update of unit magnitude. The weights affect the finite evidence offset but not this
limiting direction.

\section{Algorithm Pseudocode}
\label{app:bids-algorithm}

Algorithm~\ref{alg:bids-training} summarizes the on-policy MOPA update in
Equations~\eqref{eq:beta-gradient} and~\eqref{eq:beta-advantage}.

\begin{algorithm}[htbp]
\caption{Multi-Signal Optimization with Probabilistic Advantage (MOPA)}
\label{alg:bids-training}
\begin{algorithmic}[1]
\Require Prompt distribution $\mathcal D$; initial student parameters $\theta_0$;
fixed post-trained teacher $\pi_T$ and pre-adaptation checkpoint $\pi_{T_b}$; verifier $V$;
batch size $B$; iterations $K$; configured optimizer;
temperature $\tau>0$ and source weights $w_j>0$, $j\in\{\mathrm V,\mathrm O,\mathrm D\}$
\Ensure Trained student parameters $\theta_K$
\For{$k=0,\ldots,K-1$}
  \State Sample a batch of $B$ prompts from $\mathcal D$
  \For{each prompt $x$ in the batch}
    \State Sample $y\sim\pi_{\theta_k}(\cdot\mid x)$ with length $T$
    \State $R_{\mathrm V}\gets V(x,y)\in\{-1,1\}$
    \For{$t=1,\ldots,T$} 
      \State $h_{t}\gets(x,y_{<t})$
      \State $R_{\mathrm V,t}\gets R_{\mathrm V}$
      \State $R_{\mathrm O,t}\gets\log\pi_T(y_{t}\mid h_{t})
      -\log\pi_{\theta_k}(y_{t}\mid h_{t})$
      \State $R_{\mathrm D,t}\gets\log\pi_T(y_{t}\mid h_{t})
      -\log\pi_{T_b}(y_{t}\mid h_{t})$
      \State Compute $\widehat A_{t}$ using Equation~\eqref{eq:beta-advantage}
    \EndFor
  \EndFor
  \State Obtain $\theta_{k+1}$ with one optimizer step using Equation~\eqref{eq:beta-gradient}
\EndFor
\State \Return $\theta_K$
\end{algorithmic}
\end{algorithm}

\section{Shared Experimental Configuration}
\label{app:experiment-details}

Table~\ref{tab:shared-hyperparameters} summarizes the training configuration.
We use veRL with vLLM for rollout generation and FSDP for actor training.
Models are trained with AdamW at a constant learning rate on 32 NVIDIA
H20 GPUs, without an additional KL reward penalty for OPD.
Verifier outcomes are mapped to rewards in $\{-1,1\}$.
In multi-teacher training, each sample is routed to its domain-specific
teacher and verifier, while all samples share the same student policy.

\begin{table}[htbp]
\centering
\caption{\textbf{Shared training hyperparameters for the main distillation experiments.} Method-specific fusion parameters are reported in the main text.}
\label{tab:shared-hyperparameters}
\begin{tabular}{ll}
\hline
Hyperparameter & Value \\
\hline
Training framework & veRL \\
Rollout backend & vLLM \\
Math reward function & DAPO boxed verifier \\
Code reward function & Execution-based verifier \\
Train batch size & 1024 \\
Responses per prompt & 1 \\
PPO mini batch size & 1024 \\
PPO micro batch size/GPU & 1 \\
Max prompt length & 2048 \\
Max response length & 16384 \\
Optimizer & AdamW \\
Weight decay & 0.01 \\
Learning rate & $1\times10^{-5}$ \\
Training temperature / top-$p$ & 1.0 / 1.0 \\
Validation temperature / top-$p$ & 1.0 / 1.0 \\
Rollout tensor parallel size & 4 \\
\hline
\end{tabular}
\end{table}

Evaluation uses temperature 1.0 and top-$p$ 1.0, with a maximum generation
length of 16384 tokens. Accuracy measures sample-level correctness,
whereas pass rate measures the fraction of problems with at least one
correct response. We evaluate HumanEval+ and MBPP+ using EvalPlus with
one response per problem, and LiveCodeBench v6 with four responses per
problem.

\section{Extended Comparison with RLVR Methods}
\label{app:rlvr-results}

The main experiments follow the OPD setting, with one response sampled per
prompt. In this setting, a single binary verifier outcome provides limited
evidence for a policy update and cannot support a within-prompt relative
advantage estimate. RLVR commonly samples multiple responses per prompt and
computes group-relative advantages~\citep{deepseekmath2024}. This provides
a prompt-specific reward baseline and broader exploration, but requires more
rollout generation per prompt and still assigns credit at the response level.
We therefore extend the comparison to JustRL~\citep{he2026justrl} and
STAPO~\citep{liu2026stapo}, two RLVR methods that use only verifier feedback.
The core configuration difference is that these baselines sample eight
responses per prompt for group-based advantage estimation, whereas the
teacher-guided methods sample one and obtain dense token-level supervision
from Qwen3-4B-Math. We use the same Qwen3-4B and Qwen3-1.7B initializations,
DeepMath-103K training subset, boxed-answer verifier, and shared optimizer
settings described in Appendix~\ref{app:experiment-details}. Evaluation
follows the main protocol: 32 responses per problem for AIME24, AIME25,
and AMC23, and four for MinervaMath, with temperature and top-$p$ both 1.0.

\begin{table}[htbp]
\centering
\caption{\textbf{Mathematical reasoning results for RLVR and distillation methods, grouped by the number of training signals.} Check marks indicate the signals used for training:
$R_{\mathrm V}$ (verifier outcome), $R_{\mathrm O}$ (teacher--student
discrepancy), and $R_{\mathrm D}$ (teacher adaptation).
Benchmark scores are percentages. $\mathrm{Average}^{\dagger}$ averages
scores normalized by Qwen3-4B-Math over all four benchmarks, using this model
as a common evaluation reference for every method. Bold denotes the best result among post-training methods within each model size.}
\label{tab:rlvr-comparison}
\begingroup
\small
\setlength{\tabcolsep}{0pt}
\renewcommand{\arraystretch}{1.12}
\newcolumntype{L}[1]{>{\columncolor{white}[0pt][0.0165\linewidth]\raggedright\arraybackslash}m{#1}}
\newcolumntype{C}[1]{>{\columncolor{white}[0pt][0.0165\linewidth]\centering\arraybackslash}m{#1}}
\newcolumntype{E}[1]{>{\centering\arraybackslash}m{#1}}
\begin{tabular}{@{}L{0.155\linewidth}
  @{\hspace{0.0165\linewidth}}*{3}{C{0.032\linewidth}}
  @{\hspace{0.0165\linewidth}}*{2}{C{0.065\linewidth}}
  @{\hspace{0.0165\linewidth}}*{2}{C{0.065\linewidth}}
  @{\hspace{0.0165\linewidth}}*{2}{C{0.065\linewidth}}
  @{\hspace{0.0165\linewidth}}*{2}{C{0.065\linewidth}}
  @{\hspace{0.0165\linewidth}}C{0.065\linewidth}E{0.065\linewidth}@{}}
\toprule
\multirow{2}{*}{Method} & \multicolumn{3}{c}{Signals}
& \multicolumn{2}{c}{AIME24} & \multicolumn{2}{c}{AIME25}
& \multicolumn{2}{c}{AMC23} & \multicolumn{2}{c}{MinervaMath}
& \multicolumn{2}{c}{$\mathrm{Average}^{\dagger}$} \\
\cmidrule(lr){2-4}\cmidrule(lr){5-6}\cmidrule(lr){7-8}
\cmidrule(lr){9-10}\cmidrule(lr){11-12}\cmidrule(lr){13-14}
& $R_{\mathrm V}$ & $R_{\mathrm O}$ & $R_{\mathrm D}$
& Acc. & Pass & Acc. & Pass & Acc. & Pass & Acc. & Pass & Acc. & Pass \\
\midrule
\multicolumn{14}{l}{\textit{Reference model: Qwen3-4B-Math}} \\
Reference & -- & -- & -- & 57.40 & 76.67 & 51.67 & 66.67 & 93.52 & 97.50 & 27.85 & 33.46 & 100.00 & 100.00 \\

\midrule
\multicolumn{14}{l}{\textit{Model: Qwen3-4B}} \\
Base & -- & -- & -- & 22.60 & 60.00 & 20.83 & 33.33 & 60.16 & 92.50 & 20.31 & 27.21 & 54.24 & 76.11 \\
\addlinespace[2pt]
\multicolumn{14}{l}{\quad\textit{Single-signal}} \\
\hspace*{0.6em}+\,JustRL & $\checkmark$ & -- & -- & 53.54 & 83.33 & 42.50 & 56.67 & 88.75 & \textbf{97.50} & 25.55 & 31.99 & 90.54 & 97.32 \\
\hspace*{0.6em}+\,STAPO & $\checkmark$ & -- & -- & 55.31 & 83.33 & 50.00 & 66.67 & 90.16 & \textbf{97.50} & 27.76 & \textbf{33.46} & 97.30 & 102.17 \\
\hspace*{0.6em}+\,OPD & -- & $\checkmark$ & -- & 56.46 & 80.00 & 50.83 & 63.33 & 92.81 & 95.00 & 27.90 & 33.09 & 99.04 & 98.92 \\
\hspace*{0.6em}+\,REOPOLD & -- & $\checkmark$ & -- & 56.98 & 83.33 & 51.67 & 63.33 & 93.59 & \textbf{97.50} & 27.80 & 32.35 & 99.79 & 100.09 \\
\hspace*{0.6em}+\,AOPD & -- & $\checkmark$ & -- & 56.56 & \textbf{86.67} & 52.50 & 66.67 & 93.28 & \textbf{97.50} & 27.67 & 33.09 & 99.81 & 102.98 \\
\hspace*{0.6em}+\,TOPD & -- & $\checkmark$ & -- & 54.79 & 80.00 & 51.67 & 63.33 & 93.44 & \textbf{97.50} & 26.84 & 31.99 & 97.94 & 98.73 \\
\arrayrulecolor{black!20}\midrule[0.3pt]\arrayrulecolor{black}
\multicolumn{14}{l}{\quad\textit{Two-signal}} \\
\hspace*{0.6em}+\,ExOPD & -- & $\checkmark$ & $\checkmark$ & 57.19 & 83.33 & 52.50 & 63.33 & 93.13 & \textbf{97.50} & 28.08 & 33.27 & 100.41 & 100.78 \\
\hspace*{0.6em}+\,OWPO & $\checkmark$ & $\checkmark$ & -- & 58.33 & 80.00 & 53.85 & \textbf{76.67} & \textbf{93.75} & \textbf{97.50} & 27.76 & 31.99 & 101.44 & 103.73 \\
\hspace*{0.6em}+\,$\gamma$OPD & $\checkmark$ & $\checkmark$ & -- & 60.94 & \textbf{86.67} & 54.17 & 66.67 & 92.89 & \textbf{97.50} & 27.71 & 32.90 & 102.46 & 102.84 \\
\arrayrulecolor{black!20}\midrule[0.3pt]\arrayrulecolor{black}
\multicolumn{14}{l}{\quad\textit{Three-signal}} \\
\rowcolor{cyan!8}
\hspace*{0.6em}+\,\textbf{MOPA (ours)} & $\checkmark$ & $\checkmark$ & $\checkmark$ & \textbf{61.88} & 83.33 & \textbf{54.79} & \textbf{76.67} & 93.62 & \textbf{97.50} & \textbf{28.49} & 31.99 & \textbf{104.06} & \textbf{104.82} \\
\midrule
\multicolumn{14}{l}{\textit{Model: Qwen3-1.7B}} \\
Base & -- & -- & -- & 11.77 & 46.67 & 8.33 & 20.00 & 39.84 & 90.00 & 13.88 & 22.06 & 32.27 & 62.28 \\
\addlinespace[2pt]
\multicolumn{14}{l}{\quad\textit{Single-signal}} \\
\hspace*{0.6em}+\,JustRL & $\checkmark$ & -- & -- & 37.60 & 66.67 & \textbf{36.67} & \textbf{50.00} & 78.67 & \textbf{95.00} & 23.99 & 30.88 & 76.68 & 87.92 \\
\hspace*{0.6em}+\,STAPO & $\checkmark$ & -- & -- & 35.52 & 63.33 & 30.83 & 46.67 & 77.73 & \textbf{95.00} & 22.70 & 29.78 & 71.54 & 84.76 \\
\hspace*{0.6em}+\,OPD & -- & $\checkmark$ & -- & 37.71 & 66.67 & 31.25 & 40.00 & 76.02 & 92.50 & 23.16 & 29.78 & 72.66 & 82.71 \\
\hspace*{0.6em}+\,REOPOLD & -- & $\checkmark$ & -- & 37.50 & 66.67 & 30.83 & 43.33 & 77.73 & 92.50 & 22.15 & 29.23 & 71.91 & 83.54 \\
\hspace*{0.6em}+\,AOPD & -- & $\checkmark$ & -- & 37.71 & 70.00 & 32.50 & 43.33 & 79.38 & \textbf{95.00} & 22.15 & 28.31 & 73.25 & 84.58 \\
\hspace*{0.6em}+\,TOPD & -- & $\checkmark$ & -- & 39.38 & 70.00 & 30.83 & 36.67 & 79.14 & \textbf{95.00} & 23.25 & 27.94 & 74.10 & 81.81 \\
\arrayrulecolor{black!20}\midrule[0.3pt]\arrayrulecolor{black}
\multicolumn{14}{l}{\quad\textit{Two-signal}} \\
\hspace*{0.6em}+\,ExOPD & -- & $\checkmark$ & $\checkmark$ & 38.54 & 66.67 & 31.67 & 43.33 & 76.72 & 92.50 & 23.94 & 29.23 & 74.11 & 83.54 \\
\hspace*{0.6em}+\,OWPO & $\checkmark$ & $\checkmark$ & -- & 41.67 & 70.00 & 30.83 & 46.67 & 80.47 & \textbf{95.00} & 23.99 & 30.15 & 76.11 & 87.21 \\
\hspace*{0.6em}+\,$\gamma$OPD & $\checkmark$ & $\checkmark$ & -- & \textbf{42.08} & 70.00 & 35.00 & 46.67 & \textbf{80.86} & \textbf{95.00} & 23.98 & 30.14 & 78.40 & 87.20 \\
\arrayrulecolor{black!20}\midrule[0.3pt]\arrayrulecolor{black}
\multicolumn{14}{l}{\quad\textit{Three-signal}} \\
\rowcolor{cyan!8}
\hspace*{0.6em}+\,\textbf{MOPA (ours)} & $\checkmark$ & $\checkmark$ & $\checkmark$ & 41.73 & \textbf{73.33} & 34.12 & \textbf{50.00} & 80.45 & \textbf{95.00} & \textbf{24.99} & \textbf{31.51} & \textbf{78.62} & \textbf{90.56} \\
\bottomrule
\end{tabular}
\endgroup
\end{table}

For Qwen3-4B, Table~\ref{tab:rlvr-comparison} shows that MOPA improves
normalized average accuracy and pass rate over STAPO, the stronger RLVR
baseline, by 6.76 and 2.65 points, respectively, corresponding to relative
gains of 6.9\% and 2.6\%. The accuracy improvement extends across all four
benchmarks. The pass-rate gain is concentrated on AIME25, where a
10.00-point improvement outweighs a 1.47-point decrease on MinervaMath,
with no change on AIME24 or AMC23. MOPA also achieves the best aggregate
scores among the teacher-guided methods, indicating that its advantage
extends beyond the comparison with verifier-only training.

For Qwen3-1.7B, JustRL is the stronger RLVR baseline. MOPA improves its
normalized average accuracy by 1.94 points and pass rate by 2.64 points,
corresponding to relative gains of 2.5\% and 3.0\%. Accuracy improves on
AIME24, AMC23, and MinervaMath, offsetting a 2.55-point disadvantage on
AIME25. Pass rate increases on AIME24 and MinervaMath and is maintained
on the other two benchmarks. Although the accuracy margin is smaller than
in the 4B setting, MOPA retains the highest aggregate accuracy and pass
rate across all compared methods under the tighter capacity constraint.

These results show that group-based verifier-only training is a competitive
alternative to OPD: STAPO exceeds OPD in aggregate pass rate at 4B, while
JustRL exceeds it in both aggregate metrics at 1.7B. MOPA achieves further
gains by combining verifier feedback with student-aware teacher guidance
and the teacher-adaptation signal. Across both model sizes, this three-signal
combination yields the strongest aggregate performance among the single-,
two-, and three-signal methods, while using one training response per prompt.

\section{Computational Cost of Multi-Signal Inference}
\label{app:computational-cost}

We profile the 4B-student/4B-teacher MOPA run over all 168 recorded
training steps. Table~\ref{tab:bids-runtime-breakdown} reports wall-clock
time measured by the trainer. The step timer excludes periodic validation
and checkpointing. Rollout generation and the actor update account for
$50.31\%$ and $33.76\%$ of a training step, respectively, and together
dominate the runtime. By contrast, the complete advantage stage, which
contains both the standard advantage computation and MOPA post-processing,
accounts for only $1.48\%$.

\begin{table}[htbp]
\centering
\caption{\textbf{Per-step runtime breakdown for MOPA under same-size distillation.} }
\label{tab:bids-runtime-breakdown}
\begin{tabular}{lrr}
\hline
Operation & Time (s/step) & Step time (\%) \\
\hline
Rollout generation & 311.11 & 50.47 \\
Actor parameter update & 207.11 & 33.60 \\
Student log-probability evaluation & 29.81 & 4.84 \\
Teacher log-probability evaluation & 29.18 & 4.73 \\
Base-reference log-probability evaluation & 29.22 & 4.74 \\
Advantage and multi-signal processing & 9.10 & 1.48 \\
Verifier reward & 0.79 & 0.13 \\
Other trainer overhead & 0.16 & 0.03 \\
\hline
Total training step & 616.47 & 100.00 \\
\hline
\end{tabular}
\end{table}

The timer granularity gives a conservative upper bound on the fusion
cost. In particular, the 9.10-second advantage stage includes rollout-correction bookkeeping and metric aggregation. Therefore, the MOPA evidence transformation itself costs strictly less
than $1.48\%$ of the step time. Relative to ordinary OPD, the teacher
log-probability and student log-probability passes are already required.
Even if we conservatively charge the entire advantage stage, the verifier
reward, and the additional base-reference pass to MOPA, their combined
cost is 39.11 seconds per step, or $6.34\%$ of total training time.

As a contextual check, the raw weighted-mean three-signal run records
14.21 seconds in the same inclusive advantage timer, corresponding to
$2.02\%$ of its 703.19-second mean step time. That run was executed on a
different machine class, so the absolute times should not be interpreted
as a controlled speed comparison. Both measurements nevertheless show
that multi-signal processing is a small part of end-to-end training cost;
model generation and parameter optimization remain the principal
computational bottlenecks.

\end{document}